\pdfoutput=1
\documentclass[letterpaper]{article} % DO NOT CHANGE THIS
\usepackage[preprint]{aaai2027}  % DO NOT CHANGE THIS
\usepackage[hyphens]{url}  % DO NOT CHANGE THIS
\usepackage{graphicx} % DO NOT CHANGE THIS
\usepackage{natbib}  % DO NOT CHANGE THIS AND DO NOT ADD ANY OPTIONS TO IT
\usepackage{caption} % DO NOT CHANGE THIS AND DO NOT ADD ANY OPTIONS TO IT
\usepackage{algorithm}
\usepackage{algorithmic}
\usepackage{amsmath,amssymb}
\usepackage{booktabs}
\usepackage{newfloat}
\usepackage{listings}
\DeclareCaptionStyle{ruled}{labelfont=normalfont,labelsep=colon,strut=off} % DO NOT CHANGE THIS
\floatstyle{ruled}
\newfloat{listing}{tb}{lst}{}
\floatname{listing}{Listing}
\usepackage{fontawesome5}
\makeatletter
\newcommand{\correspondingauthor}{%
  \ifx\thanks\relax
    \textsuperscript{\faEnvelope[regular]}%
  \else
    \@ifundefined{aaai@envused}{%
      \gdef\aaai@envused{1}%
      \def\thefootnote{\faEnvelope[regular]}%
      \thanks{Corresponding authors: J.~Xu (xujing@hust.edu.cn) and X.~Wang (xgwang@hust.edu.cn).}%
    }{%
      \textsuperscript{\faEnvelope[regular]}%
    }%
  \fi
}
\makeatother

\title{ForeDrive: Foresight-Guided End-to-End Autonomous Driving \\with a Planning-Relevant Latent World Model}
\author{
    Sinuo Wang\textsuperscript{\rm 1,\rm 2}\protect\thanks{Equal contribution: S.~Wang (sinuo@hust.edu.cn) and Z.~Gu.},
    Zichong Gu\textsuperscript{\rm 2}\protect\footnotemark[1],
    Yuhan Huang\textsuperscript{\rm 1,\rm 2},
    Wenxin Wen\textsuperscript{\rm 2,\rm 3},
    Xun Yang\textsuperscript{\rm 2,\rm 3},
    Yiqing Zhang\textsuperscript{\rm 2},
    Xingyu Zhang\textsuperscript{\rm 2},
    Ningyu Che\textsuperscript{\rm 2}\protect\thanks{Project lead: N.~Che (cheningyu950@hellobike.com).},
    Jie Ling\textsuperscript{\rm 2},
    Qiankun Yu\textsuperscript{\rm 2},
    Wei Liu\textsuperscript{\rm 1},
    Jing Xu\textsuperscript{\rm 1}\correspondingauthor,
    Xinggang Wang\textsuperscript{\rm 1}\correspondingauthor
}

\affiliations{
    \textsuperscript{\rm 1}Huazhong University of Science and Technology\\
    \textsuperscript{\rm 2}Shanghai Zaofu Intelligent Technology Co., Ltd.\\
    \textsuperscript{\rm 3}Tongji University
}

\begin{document}

\maketitle

\begin{abstract}
Existing latent world models are typically optimized for future predictability, yet the resulting representations are not necessarily useful for planning in autonomous driving. Predictions are commonly used for pretraining or auxiliary supervision rather than as direct conditioning signals for trajectory generation. We propose ForeDrive, which learns a planning-relevant latent representation and couples it asymmetrically to a Diffusion Transformer (DiT) planner. The planner consumes multi-horizon latent future representations learned with a JEPA-style world model; planning gradients update the shared online encoder, while stop-gradient routing trains the latent predictor with forecasting losses only. Because predicted futures have varying reliability across horizons and BEV trajectories are misaligned with image tokens, we use gated visual fusion, future-status injection, and Trajectory-Adaptive Bias (TAB) to inject future latents as guidance without overriding the current observation. Trained with pure imitation learning and using only the current front-view image as visual input at inference, ForeDrive attains 89.9 PDMS on NAVSIM v1 and 90.0 one-stage EPDMS on NAVSIM v2, without reinforcement learning or an external trajectory scorer.
\end{abstract}

\section{Introduction}

% ¶1 Problem: why existing paradigms are insufficient
End-to-end driving planners must both anticipate how a scene may evolve and generate a trajectory that covers multiple maneuvers. Latent world models~\citep{assran2025vjepa2,zhou2025dinowm,karypidis2025dinoforsight,baldassarre2025dinofoundationvideo,wang2026drivejepa} predict future latent representations without RGB reconstruction, enabling efficient semantic foresight. Diffusion Transformer (DiT) planners such as DiffusionDrive~\citep{liao2025diffusiondrive} generate multimodal trajectories under imitation learning. Forecast accuracy, however, does not guarantee planning usefulness, as predicted latents may omit decision-critical information. Conversely, current-conditioned planners generate trajectories without explicit future representations.

\begin{figure}[t]
  \centering
  \includegraphics[width=\columnwidth]{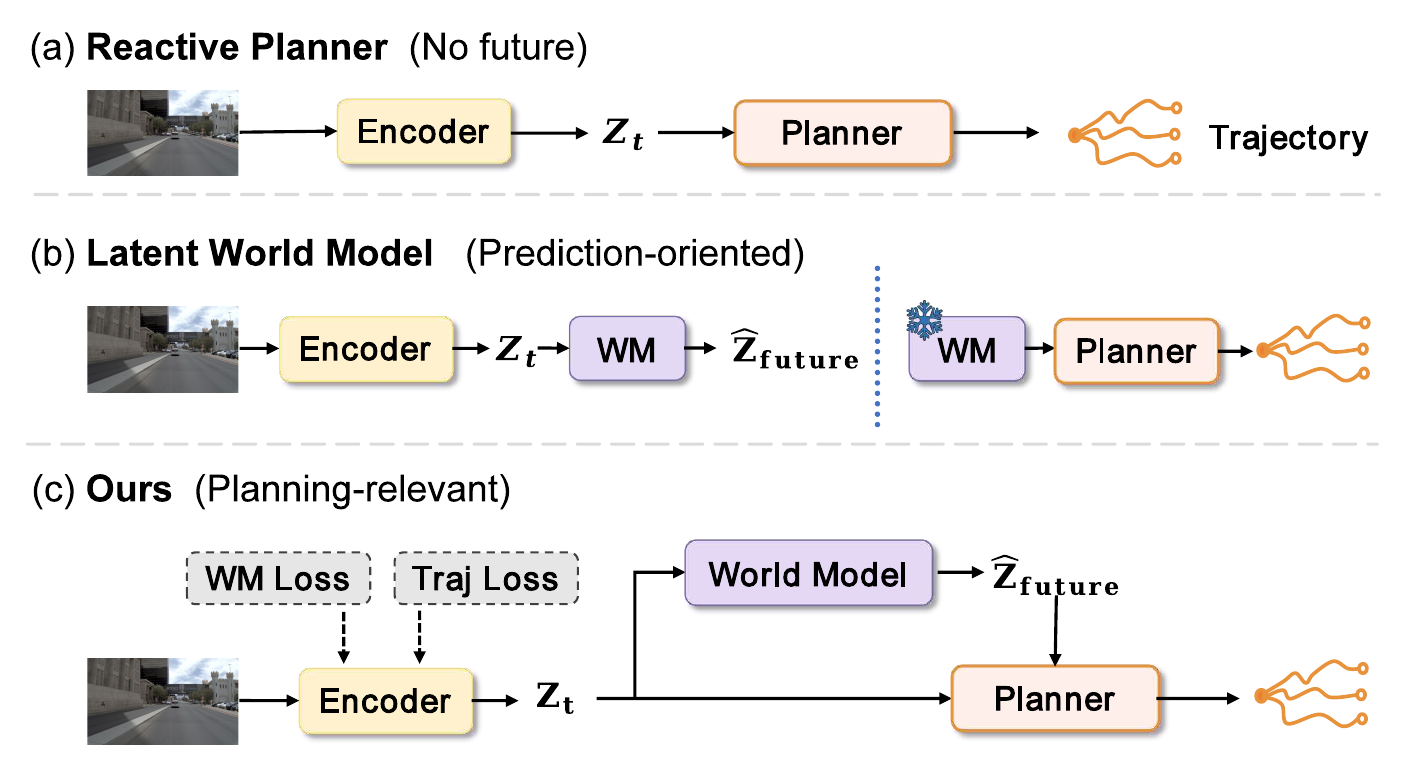}
  \caption{Representation paradigms for end-to-end planning. (a)~Reactive Planner: action generation from the current latent only (e.g., DiffusionDrive, MeanFuser). (b)~Latent World Model: predicted futures mainly for pretraining or auxiliary losses (e.g., LAW, Drive-JEPA). (c)~Ours: joint WM and trajectory supervision shapes the shared encoder, and futures guide the planner.}
  \label{fig:motivation}
\end{figure}

% ¶2 Methodological gap: why existing attempts still fail
In existing driving systems, predicted future latent representations often provide pretraining signals, auxiliary losses, or features for candidate evaluation rather than direct conditioning signals for trajectory generation~\citep{li2025law,zheng2025world4drive}. Methods that do condition planning on predicted futures commonly introduce structured scene prediction, pixel-level generation, or staged optimization~\citep{li2025bevworldmodel,zhang2025epona,xia2026drivelaw}. Few methods jointly learn a latent that is useful for planning and couple it to a generative planner while preventing planning gradients from directly rewriting the predictor (Figure~\ref{fig:motivation}).

% Flush Fig.~1 before Fig.~2 enters the queue (avoids page-1 Overfull \vbox).

% ¶3 Method (what we do; mechanism details deferred to Method / contributions)
ForeDrive jointly learns planning-relevant latent representations and integrates multi-horizon future latents as complementary multi-scale future context into a DiT planner through asymmetric latent optimization and current-anchored fusion. Because predicted futures have varying reliability across horizons, ForeDrive treats them as complementary guidance anchored by reliable current observations, rather than allowing futures to dominate current perception. A JEPA-style online/EMA predictor estimates multi-horizon visual and ego-state latents without pixel reconstruction or a separate training stage.

% ¶4 Experiments
We evaluate ForeDrive on NAVSIM under a camera-only, pure imitation-learning protocol. It attains 89.9 PDMS on v1 and 90.0 one-stage EPDMS on v2, outperforming recent end-to-end (E2E) and world-model planners under the same protocol. A zero-shot transfer to nuScenes tests cross-dataset generalization. Ablations attribute the gains mainly to future consumption, current-primary fusion, and asymmetric encoder updates.

% ¶5 Contributions
Our main contributions are summarized as follows:
\begin{itemize}
    \item We propose ForeDrive, which learns a planning-relevant latent representation and asymmetrically couples multi-horizon latent prediction with trajectory generation. Planning updates the shared encoder while stop-gradient routing prevents planning gradients from updating the latent predictor, reducing prediction--planning gradient interference.
    \item We introduce planning-oriented interfaces, including gated visual fusion, future-status injection, and Trajectory-Adaptive Bias (TAB), that incorporate predicted future dynamics into diffusion planning. These designs keep the current observation as primary evidence, and TAB links each trajectory candidate to the image tokens along its projected path during denoising.
    \item We validate ForeDrive on NAVSIM v1 and v2 under a camera-only, pure imitation-learning protocol (89.9 PDMS / 90.0 EPDMS). Using only a single front-view image at inference, ForeDrive establishes a new state-of-the-art among imitation-learning methods.
\end{itemize}

\section{Related Work}

\begin{figure}[t]
  \centering
  \includegraphics[width=\columnwidth]{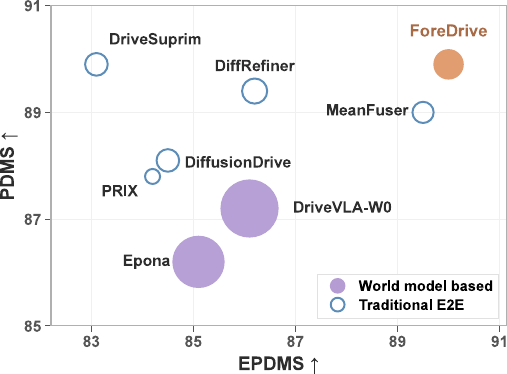}
  \caption{NAVSIM performance comparison under different model scales (EPDMS vs.\ PDMS). Only methods with reported total model parameters and both scores are shown. Marker area indicates model parameter scale (area $\propto\log$ parameter count).}
  \label{fig:data-scale}
\end{figure}

\textbf{End-to-End Autonomous Driving.}
End-to-end driving maps sensor observations directly to planned trajectories. Early systems such as TransFuser and UniAD emphasize multi-sensor fusion and BEV-centric perception--planning~\citep{chitta2023transfuser,hu2023planningoriented,li2022bevformer}, while recent work increasingly adopts camera-only inputs~\citep{liao2025diffusiondrive,wang2026meanfuser,wozniak2026prix}. Beyond imitation learning, several high-scoring methods further apply reinforcement-learning post-training, as in ReCogDrive-RL~\citep{li2026recogdrive}, or rule-based candidate scoring, as in Hydra-MDP and DriveSuprim~\citep{li2024hydramdp,yao2026drivesuprim}. These stages raise benchmark scores, but the gains of a pure imitation-learning planner without RL or external scorers remain less clear. ForeDrive therefore adopts a camera-only, pure-IL setting and examines whether a planning-relevant foresight representation improves the sensor-to-plan model without RL post-training or auxiliary scorers.

\textbf{Diffusion-based Planning.}
Diffusion models are widely used for multimodal trajectory generation in end-to-end driving, as iterative denoising can represent multiple futures and trajectory uncertainty under imitation learning. DiffusionDrive combines truncated diffusion with trajectory anchors; DiffRefiner~\citep{yin2026diffrefiner}, MeanFuser, and GoalFlow~\citep{xing2025goalflow} further develop coarse-to-fine, one-step, and flow-matching variants. These planners are typically conditioned on current or short-history features and do not condition on an explicit predicted future for planning. ForeDrive retains anchor-based diffusion decoding and feeds multi-horizon latent predictions as complementary guidance under asymmetric coupling.

\textbf{World Models for Driving.}
Prior work couples foresight and driving in four ways. (i)~Predictive world models with structured scene forecasting use BEV or occupancy futures for planning~\citep{hu2021fiery,wang2024drivingfuture,zheng2024occworld,chen2025drivinggpt,li2025bevworldmodel,zheng2025world4drive}. (ii)~JEPA-style latent predictors such as Drive-JEPA and LAW~\citep{wang2026drivejepa,li2025law} mainly treat predicted latents as pretraining or auxiliary signals rather than as inputs to a generative planner. (iii)~Video-prediction approaches such as Epona and DriveLaW~\citep{zhang2025epona,xia2026drivelaw} condition trajectory DiTs on generated video features, but rely on pixel generation and, for DriveLaW, multi-stage freezing. (iv)~Unconstrained planning-conditioned prediction allows planning objectives to reshape the foresight module without isolating forecast supervision~\citep{wang2026latentwam,li2026drivevlaw0,zhao2025policyworldmodel}.  ForeDrive instead learns planning-relevant latents and couples them to a generative planner under asymmetric optimization, keeping current evidence primary.

\section{Method}

\begin{figure*}[t]
  \centering
  \includegraphics[width=1\textwidth]{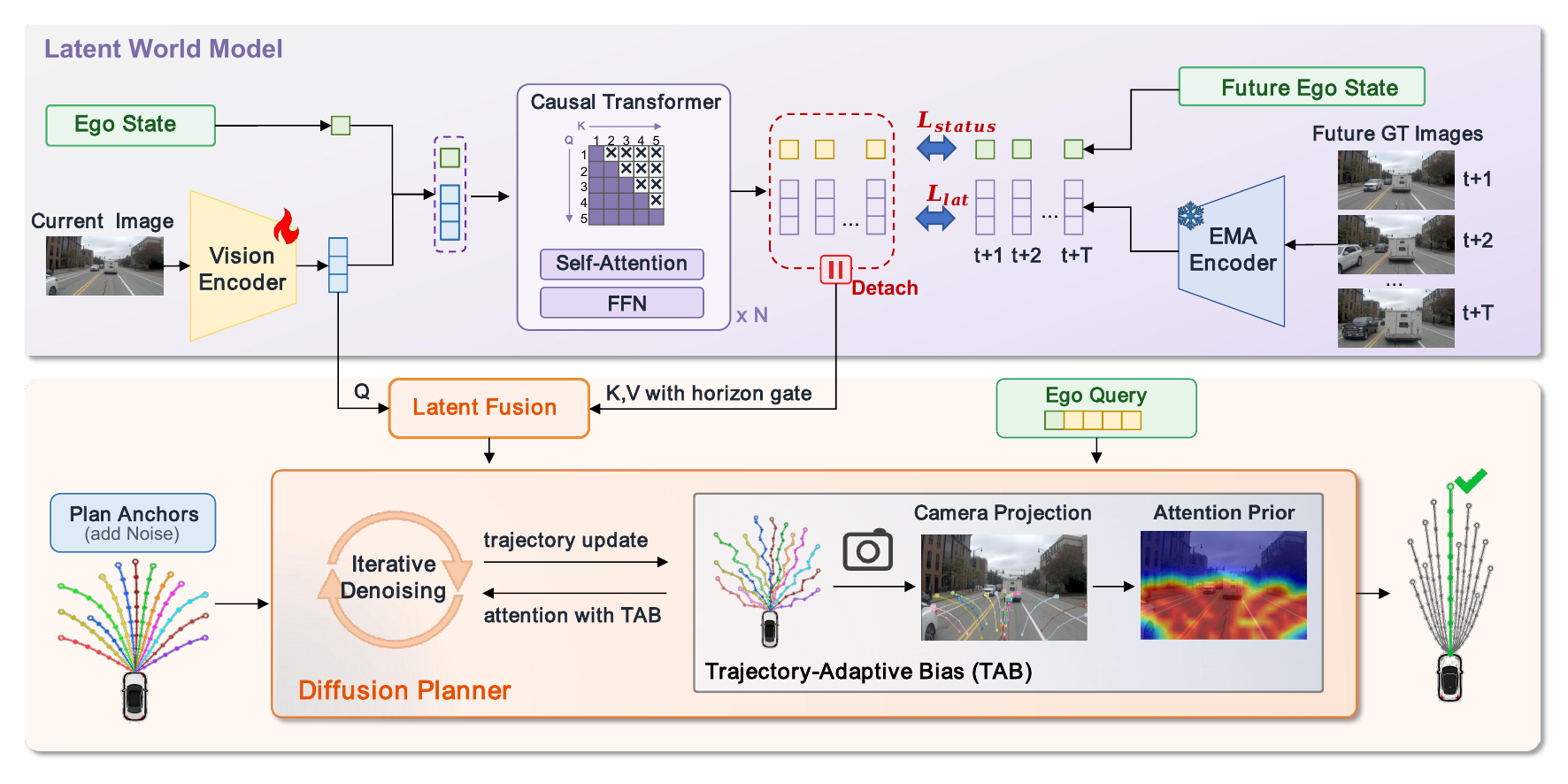} 
  \caption{Overview of ForeDrive. Multi-horizon future latents guide an anchor-based DiT via gated fusion, future-status injection, and TAB; stop-gradient routing updates the shared encoder while isolating the predictor.}
  \label{fig:overview}
  \end{figure*}

In this section, we present ForeDrive (Figure~\ref{fig:overview}). We define a planning-relevant latent as a future representation that is grounded by predictive supervision while retaining information useful for downstream trajectory generation. These objectives introduce a trade-off because forecasting favors target alignment, while planning benefits from decision-sensitive information. ForeDrive mitigates this trade-off by learning such latents with a JEPA-style world model, injecting them into a DiT planner as complementary guidance through planning-oriented interfaces (gated fusion, future-status injection, and TAB), and applying asymmetric latent optimization. Implementation details, architectural configurations, and hyperparameters are provided in the supplementary material.

\begin{table*}[t]
  \centering
  \setlength{\tabcolsep}{3.5pt}
  \begin{tabular*}{\textwidth}{@{\extracolsep{\fill}}lllcccccc}
  \toprule
  Method & Venue & Mod. & NC$\uparrow$ & DAC$\uparrow$ & TTC$\uparrow$ & Comf.$\uparrow$ & EP$\uparrow$ & PDMS$\uparrow$ \\
  \midrule
  \multicolumn{9}{@{}l}{\textit{Traditional end-to-end methods}} \\
  UniAD~\citep{hu2023planningoriented} & CVPR'23 & C & 97.8 & 91.9 & 92.9 & \textbf{100.0} & 78.8 & 83.4 \\
  TransFuser~\citep{chitta2023transfuser} & TPAMI'23 & C+L & 97.7 & 92.8 & 92.8 & \textbf{100.0} & 79.2 & 84.0 \\
  PARA-Drive~\citep{weng2024paradrive} & CVPR'24 & C & 97.9 & 92.4 & 93.0 & 99.8 & 79.3 & 84.0 \\
  DRAMA~\citep{yuan2024drama} & ISRR'24 & C+L & 98.0 & 93.1 & 94.8 & \textbf{100.0} & 80.1 & 85.5 \\
  ReCogDrive-IL~\citep{li2026recogdrive} & ICLR'26 & C & 98.1 & 94.7 & 94.2 & \textbf{100.0} & 80.9 & 86.5 \\
  PRIX~\citep{wozniak2026prix} & RA-L'26 & C & 98.1 & 96.3 & 94.1 & \textbf{100.0} & 82.3 & 87.8 \\
  DiffusionDrive~\citep{liao2025diffusiondrive} & CVPR'25 & C+L & 98.2 & 96.2 & 94.7 & \textbf{100.0} & 82.2 & 88.1 \\
  MeanFuser~\citep{wang2026meanfuser} & CVPR'26 & C & 98.6 & 97.0 & 95.0 & \textbf{100.0} & 82.8 & 89.0 \\
  DiffRefiner-R34~\citep{yin2026diffrefiner} & AAAI'26 & C & 98.4 & 97.4 & 95.3 & \textbf{100.0} & 83.4 & 89.4 \\
  \midrule
  \multicolumn{9}{@{}l}{\textit{World-model and video--action methods}} \\
  LAW~\citep{li2025law} & ICLR'25 & C & 96.4 & 95.4 & 88.7 & 99.9 & 81.7 & 84.6 \\
  Epona~\citep{zhang2025epona} & ICCV'25 & C & 97.9 & 95.1 & 93.8 & 99.9 & 80.4 & 86.2 \\
  DriveVLA-W0~\citep{li2026drivevlaw0} & ICLR'26 & C & 98.4 & 95.3 & 95.2 & \textbf{100.0} & 80.9 & 87.2 \\
  PWM~\citep{zhao2025policyworldmodel} & NeurIPS'25 & C & 98.6 & 95.9 & 95.4 & \textbf{100.0} & 81.8 & 88.1 \\
  WoTE~\citep{li2025bevworldmodel} & ICCV'25 & C+L & 98.5 & 96.8 & 94.9 & 99.9 & 81.9 & 88.3 \\
  DriveLaW~\citep{xia2026drivelaw} & CVPR'26 & C & \textbf{99.0} & 97.1 & \textbf{96.7} & \textbf{100.0} & 81.3 & 89.1 \\
  \midrule
  \textbf{ForeDrive (ours)} & --- & C & 98.6 & 97.6 & 95.6 & \textbf{100.0} & 83.8 & 89.9 \\
  ForeDrive (ViT-L) & --- & C & 98.7 & \textbf{97.9} & 96.1 & \textbf{100.0} & \textbf{84.3} & \textbf{90.4} \\
  \bottomrule
  \end{tabular*}
  \caption{Comparison on NAVSIM v1 navtest. C and C+L denote camera and camera+LiDAR. ForeDrive (ours) uses DINOv3 ViT-B/16; ForeDrive (ViT-L) is a larger-encoder upper bound.}
  \label{tab:navsim-v1}
\end{table*}

\subsection{Planning-Relevant Latent World Model}

Planning depends on future agent motion, ego-state evolution, and other scene changes that need not be represented at pixel level. This module therefore predicts future latent representations at multiple temporal scales for planning. We adopt a JEPA-style online/EMA architecture to predict latents in a DINOv3-initialized space~\citep{oquab2024dinov2,simeoni2025dinov3}. The online encoder is shared with the planner, while the predictor is optimized only by the forecasting losses defined below. This separation allows planning to shape the source representation without directly updating the predictor with planning gradients.

\paragraph{Online/EMA encoding.}
Given a current front-camera image \(I_0\) and ego status \(s_0\), an online encoder \(E_\theta\) maps \(I_0\) to \(N\) patch tokens \(z_0=E_\theta(I_0)\in\mathbb{R}^{N\times d}\). These tokens are later shared by foresight and planning. An EMA target encoder \(E_{\bar\theta}\), which copies only the vision encoder, encodes future images \(I_t\) into stop-gradient EMA targets. Future images are used only to construct training targets and are unavailable at inference.

\paragraph{Causal latent prediction.}
The current observation is represented by an ego-status token and \(N\) visual tokens, while each future horizon \(t\in\mathcal H=\{1,2,3,4\}\,\mathrm{s}\) is assigned learned query tokens, collectively denoted by \(Q_{\mathcal H}\) with \(H=|\mathcal H|\). The core prediction is
\begin{equation}
\bigl\{(\hat z_t,\hat s_t)\bigr\}_{t\in\mathcal H}
=P_\psi(z_0,s_0,Q_{\mathcal H}).
\end{equation}
All tokens are projected to a common predictor width and augmented with token-type and horizon embeddings. The Transformer predictor \(P_\psi\) processes the resulting sequence under a \emph{frame-level} block-causal mask in a single forward pass rather than an autoregressive rollout. Earlier horizons may influence later ones, while later horizons remain invisible to earlier ones. Different horizons provide complementary multi-scale future context rather than sequential rollout states. Status and visual futures are read from their corresponding query tokens.

\paragraph{Foresight supervision.}
We supervise horizon-weighted latent regression and status prediction
\begin{align}
\mathcal L_{\rm lat}
  &=\frac1H\sum_{t\in\mathcal H}w_t\,\ell_t,\\
\mathcal L_{\rm status}
  &=\frac1H\sum_{t\in\mathcal H}\bigl[
\operatorname{CE}(\hat s_t^{c},s_t^{c})
+\operatorname{MSE}(\hat s_t^{m},s_t^{m})
\bigr].
\end{align}
where \(\ell_t\) is a token-averaged \(L_1\) between predicted and EMA visual latents, \(w_t\) are fixed horizon loss weights (uniform if weighting is disabled), and \(s_t^{c}\)/\(s_t^{m}\) are the navigation-command and ego-motion (velocity/acceleration) status components.

\subsection{Foresight-Guided Generative Planning}

Because current observations are more reliable than predicted futures, ForeDrive keeps current visual tokens on the residual path and admits future latents through gated visual fusion and future-status memory. A coarse-to-fine DiT decodes multimodal trajectories from these conditions. Trajectory-Adaptive Bias (TAB) connects trajectory anchors to image tokens by projecting evolving candidates into the front view and biasing cross-attention toward path-relevant visual tokens during denoising.

\paragraph{Current-primary Latent Fusion.}
Projection necks map current tokens \(z_0\) and predicted future latent representations \(\hat z_t\) to planner-width tokens \(\bar c\) and \(\bar f_t\), respectively. We add shared spatial and per-horizon temporal embeddings, then apply a sample-shared gate \(g_t^{\rm vis}=\sigma(a_t^{\rm vis})\) as \(\tilde f_t=g_t^{\rm vis}\bar f_t\). The gate learns horizon-level contribution weights rather than sample-specific uncertainty estimates. Latent fusion \(\mathcal F\) performs residual cross-attention with \(\bar c\) as queries and concatenated gated futures as keys/values,
\begin{equation}
K,V=\operatorname{Concat}_{t\in\mathcal H}(\tilde f_t),\qquad
c^{\rm fuse}=\mathcal F(\bar c,K,V),
\end{equation}
keeping current evidence as the residual backbone.

\paragraph{Future-status injection.}
Predicted future ego motion provides complementary conditioning for trajectory generation. We embed \(\operatorname{sg}(\hat s_t^{m})\) and modulate it with an independent sample-shared gate \(g_t^{\rm st}=\sigma(a_t^{\rm st})\), yielding gated embeddings \(\tilde e_t\). These embeddings are concatenated with the current ego embedding \(e_0\) from \(s_0\) to form the planner memory
\begin{equation}
E_{\rm ego}=[e_0;\tilde e_1;\ldots;\tilde e_H],
\end{equation}
which conditions every decoder layer. Navigation commands \(\hat s_t^{c}\) are not injected through this pathway.

\begin{table*}[t]
  \centering
  \setlength{\tabcolsep}{3.5pt}
  \begin{tabular*}{\textwidth}{@{\extracolsep{\fill}}llcccccccccc}
  \toprule
  Method & Venue & NC$\uparrow$ & DAC$\uparrow$ & DDC$\uparrow$ & TLC$\uparrow$ & EP$\uparrow$ & TTC$\uparrow$ & LK$\uparrow$ & HC$\uparrow$ & EC$\uparrow$ & EPDMS$\uparrow$ \\
  \midrule
  \multicolumn{12}{@{}l}{\textit{Traditional end-to-end methods}} \\
  TransFuser~\citep{chitta2023transfuser} & TPAMI'23 & 96.9 & 89.9 & 97.8 & 99.7 & 87.1 & 95.4 & 92.7 & 98.3 & 87.2 & 76.7 \\
  ReCogDrive-IL~\citep{li2026recogdrive} & ICLR'26 & 98.2 & 94.5 & 99.3 & \textbf{99.9} & 87.4 & 97.3 & 97.1 & 98.3 & 87.2 & 86.6 \\
  PRIX~\citep{wozniak2026prix} & RA-L'26 & 98.0 & 95.6 & 99.5 & 99.8 & 87.4 & 97.2 & 97.1 & 98.3 & 87.6 & 84.2 \\
  DiffusionDrive~\citep{liao2025diffusiondrive} & CVPR'25 & 98.2 & 95.9 & 99.4 & 99.8 & 87.5 & 97.3 & 96.8 & 98.3 & 87.7 & 84.5 \\
  DiffRefiner-R34~\citep{yin2026diffrefiner} & AAAI'26 & 98.5 & 97.4 & \textbf{99.6} & 99.8 & 87.6 & 97.7 & 97.7 & 98.3 & 86.2 & 86.2 \\
  MeanFuser~\citep{wang2026meanfuser} & CVPR'26 & 98.3 & 97.2 & \textbf{99.6} & 99.8 & 87.6 & 97.4 & 97.3 & 98.3 & \textbf{88.2} & 89.5 \\
  \midrule
  \multicolumn{12}{@{}l}{\textit{World-model and video--action methods}} \\
  World4Drive~\citep{zheng2025world4drive} & ICCV'25 & 97.8 & 96.3 & 99.4 & 99.8 & 88.3 & 97.1 & 97.7 & 98.0 & 53.9 & 84.8 \\
  Epona~\citep{zhang2025epona} & ICCV'25 & 97.1 & 95.7 & 99.3 & 99.7 & \textbf{88.6} & 96.3 & 97.0 & 98.0 & 67.8 & 85.1 \\
  DriveVLA-W0~\citep{li2026drivevlaw0} & ICLR'26 & 98.5 & \textbf{99.1} & 98.0 & 99.7 & 86.4 & 98.1 & 93.2 & 97.9 & 58.9 & 86.1 \\
  WorldRFT~\citep{yang2026worldrft} & AAAI'26 & 97.8 & 96.5 & 99.5 & 99.8 & 88.5 & 97.0 & 97.4 & 98.1 & 69.1 & 86.7 \\
  Drive-JEPA~\citep{wang2026drivejepa} & arXiv'26 & 98.4 & 98.6 & 99.1 & 99.8 & 88.4 & 97.8 & 97.6 & 97.9 & 84.8 & 87.8 \\
  Latent-WAM~\citep{wang2026latentwam} & arXiv'26 & 98.1 & 97.3 & \textbf{99.6} & 99.8 & 87.7 & 97.3 & 97.6 & 98.1 & 87.3 & 89.3 \\
  \midrule
  \textbf{ForeDrive (ours)} & --- & 98.6 & 97.6 & \textbf{99.6} & 99.8 & 87.6 & 97.9 & \textbf{97.9} & 98.3 & 87.5 & 90.0 \\
  ForeDrive (ViT-L) & --- & \textbf{98.7} & 97.9 & \textbf{99.6} & \textbf{99.9} & 87.6 & \textbf{98.2} & 97.7 & \textbf{98.4} & 88.1 & \textbf{90.6} \\
  \bottomrule
  \end{tabular*}
  \caption{Comparison on NAVSIM v2 navtest (one-stage non-reactive EPDMS). ForeDrive (ours) uses DINOv3 ViT-B/16; ForeDrive (ViT-L) is a larger-encoder upper bound.}
  \label{tab:navsim-v2}
\end{table*}

\begin{table*}[t]
  \centering
  \setlength{\tabcolsep}{6pt}
  \begin{tabular}{@{}lcccccccc@{}}
  \toprule
  Method & \multicolumn{4}{c}{L2 (m)$\downarrow$} & \multicolumn{4}{c}{Collision (\%)$\downarrow$} \\
  \cmidrule(lr){2-5}\cmidrule(lr){6-9}
  & 1\,s & 2\,s & 3\,s & Avg. & 1\,s & 2\,s & 3\,s & Avg. \\
  \midrule
  PWM (FT)~\citep{zhao2025policyworldmodel}
    & 0.41 & 0.75 & 1.17 & 0.78 & 0.02 & 0.10 & 0.35 & 0.16 \\
  Epona~\citep{zhang2025epona}
    & 0.96 & 1.59 & 2.32 & 1.62 & 0.09 & 0.27 & 0.67 & 0.34 \\
  DriveLaW~\citep{xia2026drivelaw}
    & 0.30 & 0.48 & 0.83 & 0.54 & 0.23 & 0.16 & \textbf{0.19} & 0.19 \\
  ForeDrive (ours)
    & \textbf{0.22} & \textbf{0.44} & \textbf{0.79} & \textbf{0.48}
    & \textbf{0.01} & \textbf{0.09} & 0.23 & \textbf{0.11} \\
  \bottomrule
  \end{tabular}
  \caption{Zero-shot planning performance on the nuScenes validation set under the VAD~/ST-P3 protocol. PWM (FT) uses nuScenes-trained checkpoints.}
  \label{tab:nuscenes-zero-shot}
\end{table*}

\paragraph{Coarse-to-fine diffusion planner.}
Following DiffusionDrive~\citep{liao2025diffusiondrive}, we apply truncated diffusion with a cascaded DiT decoder to a fixed set of trajectory anchors \(A\). Conditioned on fused visual tokens \(c^{\rm fuse}\) and ego memory \(E_{\rm ego}\), the planner predicts multimodal trajectories \(\tau\in\mathbb{R}^{8\times3}\) over \(4\,\mathrm{s}\). Each decoder stage uses hard closest-anchor assignment, sigmoid focal loss for mode classification, and \(L_1\) regression for the winning mode:
\begin{equation}
\mathcal L_{\rm plan}=\sum_{k=1}^{2}
\left(\lambda_{\rm cls}\mathcal L_{\rm focal}^{(k)}
+\lambda_{\rm reg}\mathcal L_1^{(k)}\right).
\end{equation}
At inference, truncated DDIM runs for two steps and selects the mode by \(\arg\max\) of the classification head.

\paragraph{Trajectory-Adaptive Bias for Planning.}
Anchor-based DiT decoding represents trajectory candidates in ego/BEV coordinates, while the visual stream consists of front-view image tokens. Standard trajectory-to-visual cross-attention therefore lacks an explicit correspondence between BEV paths and image patches. We introduce Trajectory-Adaptive Bias (TAB) as a trajectory-aware visual attention interface to connect BEV trajectory candidates with front-view image tokens. For each mode, TAB projects the current ego/BEV trajectory candidate onto the front-camera image and builds a soft Gaussian proximity field over visual tokens; its log affinity is added to the cross-attention logits, so each mode attends more strongly to visual tokens near its projected path while retaining access to the full scene.

As denoising updates the trajectory candidates, TAB recomputes the bias at stage boundaries. The bias is mode-specific and differentiable; if a candidate has no valid projection, the added bias is constant and leaves the attention distribution unchanged. We apply TAB in every decoder layer of both stages, independently of future injection.

\subsection{Asymmetric Latent Optimization}
The online representation \(z_0\) is shared by forecasting and planning, so the two objectives may introduce optimization interference. Back-propagating the planning loss through the predictor can turn \(P_\psi\) into a planning feature adapter and weaken forecast fidelity, whereas freezing the encoder for forecasting prevents planning from shaping actionable futures. ForeDrive mitigates this interference with asymmetric latent optimization via stop-gradient routing: both losses update the shared encoder, but planning gradients do not update the latent predictor.

Asymmetric gradient routing is given by
\begin{equation}
\mathcal L_{\rm plan}\;\rightarrow\; E_\theta,\;\mathcal F,\;\text{planner},
\qquad
\mathcal L_{\rm plan}\;\not\rightarrow\; P_\psi.
\end{equation}
The latent and status objectives train \(P_\psi\). Planning updates \(E_\theta\), and the EMA target encoder tracks \(E_\theta\) by exponential moving average, so planning affects EMA targets only through this encoder--EMA path. Injected future latent representations remain stop-gradient.

The full training objective is
\begin{equation}
\mathcal L
=\lambda_{\rm traj}\mathcal L_{\rm plan}
+\lambda_z\mathcal L_{\rm lat}
+\lambda_s\mathcal L_{\rm status}.
\end{equation}
All objectives are optimized in one stage, and the EMA target receives no gradient. The EMA target and auxiliary status objective help avoid representational collapse.

\section{Experiments}
\label{sec:experiments}

\subsection{Setup}

We train on the official NAVSIM \texttt{navtrain} split and report final metrics on \texttt{navtest}. The default backbone is DINOv3 ViT-B/16 with a \(256\times512\) front image; the WM predicts futures at \(1/2/3/4\,\mathrm{s}\), and the planner uses 20 trajectory anchors over a \(4\,\mathrm{s}\) horizon. Default ForeDrive uses a 4-layer predictor, a \(2{\times}5\) DiT, and AdamW at \(6{\times}10^{-4}\) for 100 epochs (batch 1024). Full configs are in the supplementary material.

\subsection{Benchmark}

We evaluate on NAVSIM v1 and v2~\citep{dauner2024navsim}. The primary metrics are PDMS on v1 and one-stage EPDMS on v2 \texttt{navtest}; we also report the associated safety and progress submetrics. NAVSIM uses \emph{non-reactive simulation}: the ego vehicle commits to one planned trajectory over a fixed horizon while surrounding agents follow log replay. We use \emph{one-stage EPDMS} for the Extended PDM Score computed from a single \(4\,\mathrm{s}\) horizon of real observations under this protocol.

\subsection{Main Results}

\paragraph{Comparison on NAVSIM v1.}
Default ForeDrive (ViT-B/16) attains 89.9 PDMS on NAVSIM v1 (Table~\ref{tab:navsim-v1}), improving over DiffRefiner-R34 and MeanFuser by 0.5 and 0.9 points, DiffusionDrive by 1.8, and DriveLaW and PWM by 0.8 and 1.8. Retraining the default recipe from three different random seeds yields PDMS in 89.7--89.9 with a standard deviation of 0.1; the table reports the primary run. Scaling only the vision encoder to ViT-L/16 further reaches 90.4 PDMS. Under the pure-IL protocol, these gains hold across seeds.

\paragraph{Comparison on NAVSIM v2.}
On NAVSIM v2, default ForeDrive (ViT-B/16) reaches 90.0 one-stage EPDMS (Table~\ref{tab:navsim-v2}), improving over MeanFuser by 0.5 points and over Latent-WAM and Drive-JEPA by 0.7 and 2.2 points, respectively. The ViT-L upper bound further attains 90.6. Together with the v1 results, the same pure-IL recipe yields consistent margins on both metric suites. Figure~\ref{fig:data-scale} compares PDMS/EPDMS across methods of different scales: at 118.4M parameters, ForeDrive matches the best listed PDMS and attains the highest EPDMS among the plotted methods, while remaining substantially smaller than Epona (2.5B) and DriveVLA-W0 (7.5B).

\begin{figure}[t]
  \centering
  \includegraphics[width=1\linewidth]{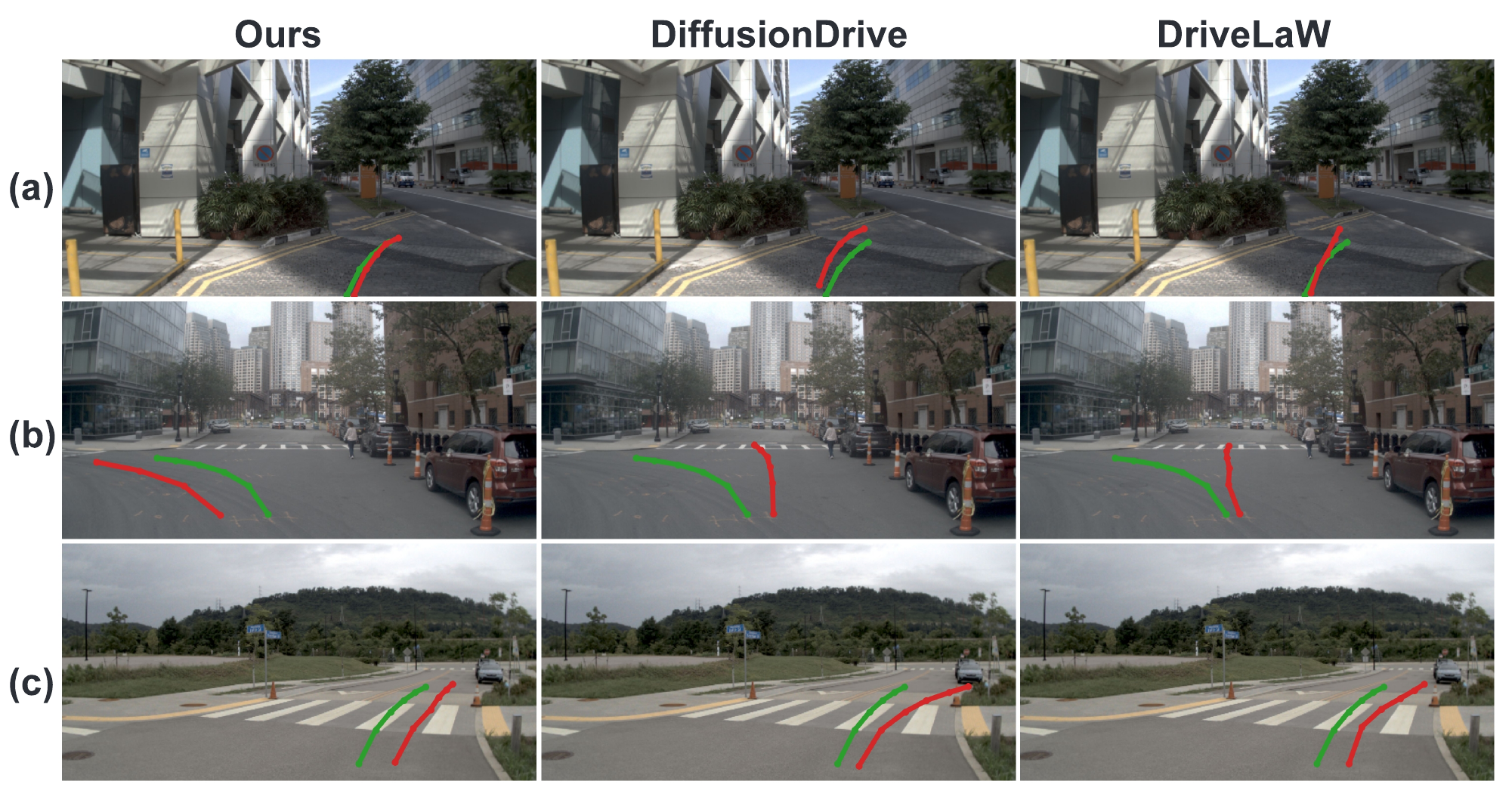}
  \caption{Trajectory comparison with DiffusionDrive and DriveLaW (green: expert; red: prediction).}
  \label{fig:bev}
\end{figure}

\paragraph{Zero-shot performance on nuScenes.}
Table~\ref{tab:nuscenes-zero-shot} reports open-loop planning on the nuScenes validation set under the VAD~/ST-P3 protocol~\citep{jiang2023vad,hu2022stp3}, with all methods re-evaluated under a shared metric implementation. ForeDrive is transferred zero-shot from NAVSIM without nuScenes fine-tuning. It attains the lowest L2 at every horizon and the lowest collision at 1\,s, 2\,s, and on average, reaching 0.48\,m / 0.11\% versus 0.54\,m / 0.19\% for DriveLaW and 0.78\,m / 0.16\% for nuScenes-trained PWM (FT); DriveLaW remains lower at collision@3\,s (0.19\% vs.\ 0.23\%). These results indicate that foresight-guided planning learned on NAVSIM transfers across datasets in both displacement accuracy and safety.

\paragraph{Qualitative results.}
Figure~\ref{fig:bev} compares ForeDrive with DiffusionDrive and DriveLaW on three interactive scenes (green: expert; red: prediction). ForeDrive stays closer to the expert corridor where the reactive DiT baselines leave the lane or incur safety/progress failures. These cases illustrate that foresight-guided planning helps on maneuvers that are difficult for current-only DiT planners.

\subsection{Ablation Study}

\paragraph{Base configuration.}
The planner retains DiffusionDrive's anchor-based truncated-diffusion formulation but replaces its camera--LiDAR BEV interface with DINOv3 features from one current front-view image. We then add FiLM conditioning on the current ego state and deepen both DiT stages from one to five layers. The resulting current-only model is \textbf{Base}. Base and ForeDrive use the same DINOv3 ViT-B/16 encoder, $2{\times}5$-layer DiT, navigation command, and positional and temporal embeddings. Base contains neither the world model (WM) nor TAB. The supplementary material reports the stepwise construction at ViT-S scale.

\begin{table}[t]
  \centering
  \setlength{\tabcolsep}{1.4mm}
  \begin{tabular}{@{}lcccc@{}}
  \toprule
  Setting & WM & TAB & PDMS$\uparrow$ & $\Delta$ vs.\ Base \\
  \midrule
  DiffusionDrive &  &  & 88.1 & -- \\
  \midrule
  Base &  &  & 88.9 & -- \\
  Base + TAB &  & \(\checkmark\) & 89.5 & \(+0.6\) \\
  Base + WM & \(\checkmark\) &  & 89.6 & \(+0.7\) \\
  Base + WM + TAB & \(\checkmark\) & \(\checkmark\) & \textbf{89.9} & \(\mathbf{+1.0}\) \\
  \bottomrule
  \end{tabular}
  \caption{Ablation study of WM and TAB relative to Base. DiffusionDrive is an external reference. Here WM denotes the complete future representation pipeline, including latent prediction supervision and future conditioning interfaces.}
  \label{tab:wm-tab-factorial}
  \end{table}

\begin{table}[t]
  \centering
  \setlength{\tabcolsep}{0.8mm}
  \begin{tabular}{@{}lccccc@{}}
  \toprule
  Setting & Aux. & \(z\) Inject. & \(s\) Inject. & PDMS$\uparrow$ & $\Delta$ vs.\ Base \\
  \midrule
  Base &  &  &  & 88.9 & -- \\
  Auxiliary & \(\checkmark\) &  &  & 89.0 & \(+0.1\) \\
  Visual only & \(\checkmark\) & \(\checkmark\) &  & 89.4 & \(+0.5\) \\
  Status only & \(\checkmark\) &  & \(\checkmark\) & 89.3 & \(+0.4\) \\
  Full WM & \(\checkmark\) & \(\checkmark\) & \(\checkmark\) & \textbf{89.6} & \(\mathbf{+0.7}\) \\
  \bottomrule
  \end{tabular}
  \caption{Ablation study of future representation components without TAB on NAVSIM v1 (auxiliary loss, visual injection, and status injection).}
  \label{tab:ablation}
  \end{table}

\begin{table}[t]
  \centering
  \setlength{\tabcolsep}{1.8pt}
  \begin{tabular}{@{}lcccc@{}}
  \toprule
  Training paradigm
  & \(\mathcal{L}_{\rm lat}\)\(\downarrow\)
  & Lat.\ Cos.\(\uparrow\)
  & \(\mathcal{L}_{\rm status}\)\(\downarrow\)
  & PDMS\(\uparrow\) \\
  \midrule
  Two-stage / freeze WM & \textbf{5.93} & \textbf{0.843} & 0.604  & 87.9 \\
  Joint + detach encoder & 6.07 & 0.838 & 0.599 & 88.6  \\
  Joint + aux-only & 7.44 & 0.769 & 0.536 & 89.6  \\
  Full joint (ours) & 7.64 & 0.759 & \textbf{0.531} & \textbf{89.9}\\
  \bottomrule
  \end{tabular}
  \caption{Comparison of training paradigms on NAVSIM. We report latent \(L_1\), latent cosine similarity (Lat.\ Cos.), composite future-status loss, and navtest PDMS.}
  \label{tab:train-paradigm-B}
  \end{table}

\begin{table}[t]
  \centering
  \setlength{\tabcolsep}{1.8mm}
  \begin{tabular}{@{}lcc@{}}
  \toprule
  Interface & PDMS$\uparrow$ & $\Delta$ vs.\ ours \\
  \midrule
  Gated fusion (ours) & \textbf{89.4} & 0.0 \\
  \midrule
  Current only & 88.9 & $-$0.5 \\
  Future only & 84.2 & $-$5.2 \\
  Concatenation & 89.2 & $-$0.2 \\
  Dual-memory & 89.0 & $-$0.4 \\
  Ungated fusion & 89.0 & $-$0.4 \\
  \bottomrule
  \end{tabular}
  \caption{Future-injection interfaces on NAVSIM (TAB and future-status injection off).}
  \label{tab:fusion-B}
  \end{table}

\begin{table}[t]
  \centering
  \setlength{\tabcolsep}{2.2mm}
  \begin{tabular}{@{}llcccc@{}}
  \toprule
  Encoder & Params & Base$\uparrow$ & Full$\uparrow$ & $\Delta$ \\
  \midrule
  DINOv3 ViT-S/16 & 21M & 87.4 & 89.0 & $+1.6$ \\
  DINOv3 ViT-B/16 & 86M & 88.9 & 89.9 & $+1.0$ \\
  DINOv3 ViT-L/16 & 300M & 89.3 & 90.4 & $+1.1$ \\
  \bottomrule
  \end{tabular}
\caption{DINOv3 backbone capacity on NAVSIM v1 under a fixed ForeDrive pipeline. Params counts the vision encoder only. Base is the matched current-only planner; Full is ForeDrive (WM+TAB); $\Delta$ is Full$-$Base. ViT-L is a capacity upper bound; main results and ablations use ViT-B.}
  \label{tab:encoder}
\end{table}

\paragraph{Component ablation.}
Table~\ref{tab:wm-tab-factorial} ablates WM and TAB on the matched Base (88.9 PDMS), with DiffusionDrive listed only as an external reference (88.1). Adding TAB or WM alone raises PDMS to 89.5 (\(+0.6\)) and 89.6 (\(+0.7\)), respectively, while combining both reaches 89.9 (\(+1.0\)). The joint gain exceeds either factor alone, indicating that WM and TAB contribute distinct effects.

Table~\ref{tab:ablation} further decomposes the WM under TAB off. Auxiliary prediction alone yields only \(+0.1\) PDMS, indicating that a forecasting side objective is insufficient. Exposing predicted futures to the planner accounts for most of the gain: visual injection reaches \(+0.5\) and future-status injection \(+0.4\); using both pathways attains \(+0.7\) and recovers the Base${+}$WM result in Table~\ref{tab:wm-tab-factorial}. The main improvement thus comes from consuming predicted futures, not from adding a forecasting loss alone.

\paragraph{Training paradigm.}
Table~\ref{tab:train-paradigm-B} compares four training paradigms along two axes: whether planning gradients update the shared online encoder, and whether predicted futures are injected. \emph{Two-stage / freeze WM} pretrains then freezes the world model; \emph{Joint + detach encoder} trains jointly but stops planner gradients before the shared encoder; \emph{Joint + aux-only} updates the encoder with both losses yet does not inject futures; \emph{Full joint} is our setting, with planning-driven encoder updates, future injection, and forecasting-only supervision of the predictor.

The first two settings best match EMA visual targets (lowest \(\mathcal{L}_{\rm lat}\) 5.93 / 6.07; highest latent cosine 0.843 / 0.838) yet obtain the weakest PDMS (87.9 / 88.6). Encoder-updating joint training raises PDMS to 89.6--89.9 despite weaker visual alignment. Under matched future injection, full joint improves over encoder detachment by 1.3 PDMS; with encoder updates retained, enabling injection adds 0.3 over the auxiliary-only joint baseline. Across variants, better generic latent forecast alignment does not necessarily correspond to higher planning scores, indicating that planning-oriented representations may deviate from prediction-optimal targets to preserve decision-relevant information.

\paragraph{Latent fusion strategies.}
Under a matched protocol with TAB and future-status ego-KV disabled, we compare future-injection interfaces in Table~\ref{tab:fusion-B}. Gated current-primary fusion attains 89.4 PDMS; current-only is lower by 0.5, while future-only drops by 5.2. Predicted futures therefore improve planning only when fused with the present observation, and current evidence should remain primary. Concatenation, dual-memory, and ungated fusion trail gated fusion by 0.2--0.4, indicating that a gated residual interface is preferable to exposing additional future tokens alone. Under the same TAB-/status-off protocol, permuting predicted horizons at inference leaves PDMS unchanged at \(89.4\), whereas zeroing futures at test time reduces it to \(88.0\). The permutation result suggests that the planner mainly exploits aggregated multi-scale future context rather than strict horizon ordering, which is consistent with our parallel latent prediction design. Full breakdowns are reported in the supplementary material.

\paragraph{Encoder capacity.}
Table~\ref{tab:encoder} varies only the DINOv3 backbone under a fixed ForeDrive pipeline. Stronger encoders raise both Base and Full: a better current representation already improves the current-only planner, and Full improves as well (89.0 / 89.9 / 90.4 at ViT-S/B/L). Full still outperforms its matched Base at every scale. The Base${\to}$Full margin is largest on ViT-S (\(+1.6\)), where perception is weakest, and remains positive on ViT-B and ViT-L (\(+1.0\) / \(+1.1\)). Foresight and TAB still help at every encoder scale; they do not replace a stronger backbone. Single-factor and WM-component breakdowns at ViT-S/L appear in the supplementary material. We keep ViT-B/16 for the main results and ablations for its accuracy--cost trade-off, and report ViT-L only as a capacity upper bound.

\paragraph{Inference efficiency.}
On a single H20 GPU, the default ForeDrive runs at \(58.2\,\mathrm{ms}\) per frame (17.2 FPS) with \(1.03\,\mathrm{GB}\) peak memory, while latent foresight adds approximately \(26\,\mathrm{ms}\) over the current-only Base. The full model has 118.4M parameters at evaluation; the training-only EMA encoder is excluded from this total.

\section{Conclusion}

ForeDrive learns planning-relevant future latent representations and couples them asymmetrically to a DiT planner. Asymmetric latent optimization via stop-gradient routing mitigates direct prediction--planning optimization interference, while planning-oriented interfaces that include gated fusion, future-status injection, and TAB connect predicted futures to diffusion planning. With a single front-view image at inference and pure imitation learning, ForeDrive attains 89.9 PDMS on NAVSIM v1 and 90.0 EPDMS on v2. Matched ablations show that auxiliary forecasting alone is insufficient, that incorporating future latents is necessary, and that encoder updates yield planning-relevant futures despite weaker EMA alignment.

The evaluation is limited to camera-only, non-reactive simulation, where inaccurate futures can still mislead the planner. Testing under interactive closed-loop settings is an important next step.

\bibliography{refs}

\end{document}

\twocolumn[{%
  \centering
  \vspace*{0.15in}
  {\LARGE\bf Supplementary Material\par}
  \vspace{0.35in}
}]

\appendix

\noindent\textit{Organization.}
We organize the appendix by question:
Sec.~\ref{app:setup} gives implementation details;
Sec.~\ref{app:base-decomp} builds the ViT-S Base for small-encoder ablations;
Sec.~\ref{app:uv-attention} visualizes TAB on the ViT-B Base;
Sec.~\ref{app:future-causal} tests causal use of predicted futures (TAB/status off);
Sec.~\ref{app:wm-compare} contrasts Full (WM+TAB+status) with Base+TAB;
Sec.~\ref{app:horizon} varies multi-scale future-query sets;
Sec.~\ref{app:encoder-ablations} repeats WM/TAB factorials at ViT-S and ViT-L;
Sec.~\ref{app:detach-vs-joint} isolates stop-gradient routing for future latent injection;
Sec.~\ref{app:traj-probe} probes trajectory readability under a frozen backbone;
Sec.~\ref{app:rgb-probe} checks scene grounding via an RGB readout;
Sec.~\ref{app:fig2-data} lists parameter counts for the main accuracy--size plot.

\section{More Implementation Details}
\label{app:setup}

We detail the data pipeline, latent world model, current-primary foresight interface, Trajectory-Adaptive Bias (TAB), truncated-diffusion planner, and joint training recipe.

\paragraph{Data pipeline and evaluation protocol.}
We train on official NAVSIM \texttt{navtrain}, with 18{,}179 held-out validation samples and 85{,}109 training samples.
Checkpoints are selected on validation; the 12{,}147-scenario \texttt{navtest} set is used only for final reporting.
Unless noted otherwise, we report NAVSIM v1 PDMS and NAVSIM v2 one-stage EPDMS under official non-reactive simulation.
The ego commits to one \(4\,\mathrm{s}\) plan while surrounding agents follow log replay.
We do not use two-stage pseudo-simulation.

The sensor input is one front-camera frame.
We crop 28 pixels from the top and bottom, resize to \(256\times512\), and apply no stochastic augmentation.
Future frames are used only to build EMA latent targets and are unavailable at inference.
By default, the world model predicts four horizons at \(1/2/3/4\,\mathrm{s}\) on the 2\,Hz NAVSIM grid.
The planning target is an eight-pose trajectory \((x,y,\theta)\) over \(4\,\mathrm{s}\) at \(0.5\,\mathrm{s}\) steps, decoded from 20 anchors.

\paragraph{Latent world model.}
The default backbone is DINOv3 ViT-B/16.
It encodes the current front image into 512 patch tokens of width 768.
An EMA copy of the vision encoder provides stop-gradient targets for future frames; only this encoder is mirrored.
A lightweight Transformer predictor forecasts multi-horizon (multi-scale) visual latents and ego-status quantities from current tokens and learnable future queries.

The predictor is a 4-layer pre-LN Transformer of width 512 (8 heads, FFN 2048, GELU, dropout 0.1).
Linear projections bridge the 768-D encoder space and the predictor width; time/token embeddings and future queries share that width.
Attention uses a frame-level block-causal mask: tokens within a horizon attend freely, while later horizons stay invisible.
All horizons are predicted in one forward pass, not by autoregressive rollout.
Each horizon emits a status token for a 4-way navigation-command classifier (CE) and raw velocity/acceleration regressions (MSE).
Predicted visual futures are projected back to 768-D before the latent loss.

Foresight supervision covers the four default horizons.
Visual terms use token-averaged \(L_1\) against EMA targets, without a patch mask.
Horizon weights are proportional to \(1/(i{+}1)\) on the 2\,Hz index \(i\), then rescaled to unit mean.
The status objective equally weights command CE and motion MSE, and is scaled by \(0.1\) in the joint loss.

\begin{table}[t]
  \centering
  \setlength{\tabcolsep}{2mm}
  \begin{tabular}{@{}clc@{}}
  \toprule
  Step & Configuration & PDMS$\uparrow$ \\
  \midrule
  Ref. & DiffusionDrive (camera+LiDAR BEV) & 88.1 \\
  \midrule
  \multicolumn{3}{@{}l}{\textit{Input/representation adaptation}} \\
  (a) & DINOv3 ViT-S camera only & 86.4 \\
  (b) & a + current-status FiLM & 86.6 \\
  (c) & b + deepen DiT ($1{\to}5$/stage)\ (\textbf{= ViT-S Base}) & 87.4 \\
  \bottomrule
  \end{tabular}
  \caption{From DiffusionDrive to ViT-S Base on NAVSIM v1 (DINOv3 ViT-S/16, camera-only). Steps~(a)--(c) are uncontrolled input/representation adaptations. DiffusionDrive is not a controlled baseline due to LiDAR input; it is listed as an external reference only.}
  \label{tab:base-decomp}
\end{table}

\begin{figure*}[t]
  \centering
  \includegraphics[width=\textwidth]{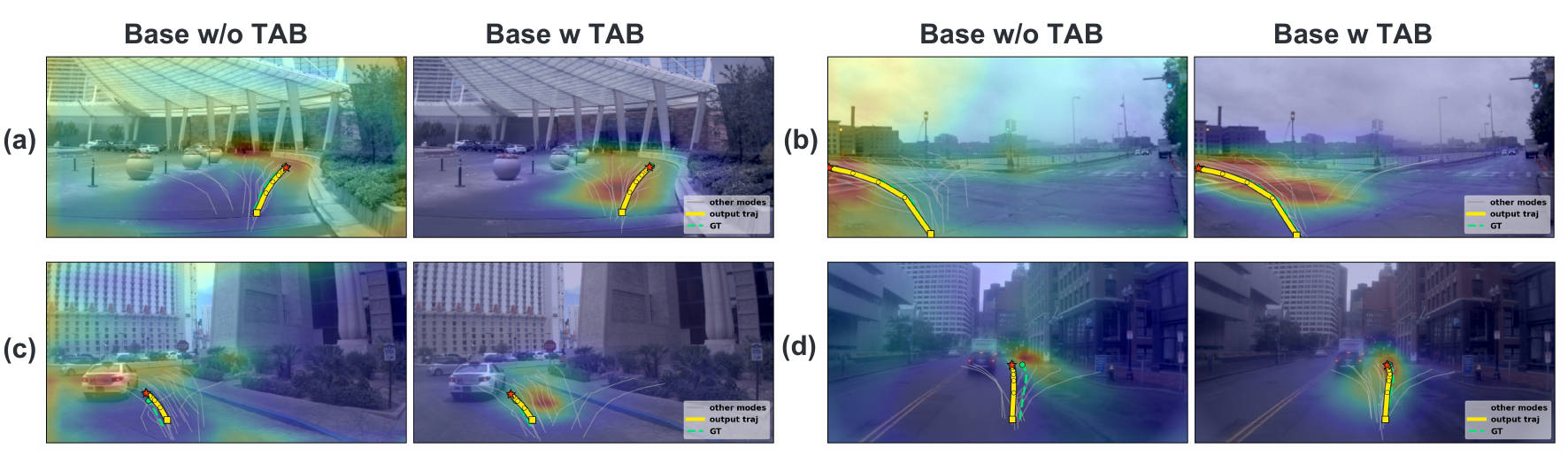}
  \caption{Trajectory-to-image attention with vs.\ without TAB (ViT-B Base; no WM). Left: Base (\(\sigma{=}0\)); right: Base+TAB (\(\sigma{=}0.25\)). Heatmaps average decoder attention over layers and trajectory-mode queries; yellow/cyan: selected plan / expert. TAB concentrates mass on the near-road corridor.}
  \label{fig:uv-attention}
\end{figure*}

\begin{table*}[t]
  \centering
  \setlength{\tabcolsep}{5pt}
  \begin{tabular*}{\textwidth}{@{\extracolsep{\fill}}lcccccccc@{}}
  \toprule
  Mode & NC & DAC & EP & TTC & C & DDC & PDMS & $\Delta$ vs.\ normal \\
  \midrule
  oracle
    & \(98.9\) & \(97.0\) & \(83.5\) & \(96.0\) & \(100.0\) & \(98.5\) & \(89.6\) & \(+0.2\) \\
  horizon\_shuffle
    & \(98.7\) & \(97.1\) & \(83.6\) & \(95.4\) & \(100.0\) & \(98.3\) & \(89.4\) & \(+0.0\) \\
  \textbf{normal}
    & \(98.7\) & \(97.1\) & \(83.6\) & \(95.4\) & \(100.0\) & \(98.3\)
    & \(\mathbf{89.4}\) & \(0\) \\
  token\_mask (\(r{=}0.75\))
    & \(98.9\) & \(96.7\) & \(82.7\) & \(95.7\) & \(100.0\) & \(98.4\) & \(89.0\) & \(-0.4\) \\
  zero
    & \(98.9\) & \(95.8\) & \(81.2\) & \(95.9\) & \(99.9\) & \(98.4\) & \(88.0\) & \(-1.4\) \\
  persistence
    & \(98.3\) & \(95.6\) & \(81.7\) & \(93.7\) & \(100.0\) & \(98.1\) & \(87.0\) & \(-2.4\) \\
  cross\_sample
    & \(98.0\) & \(95.9\) & \(81.7\) & \(93.2\) & \(99.9\) & \(98.3\) & \(87.0\) & \(-2.4\) \\
  \bottomrule
  \end{tabular*}
  \caption{Causal interventions on future latents without TAB/status (gated fusion only). $\Delta$: PDMS vs.\ normal. Subscore abbreviations follow NAVSIM (NC/DAC/EP/TTC/C/DDC).}
  \label{tab:app-causal-no-crutch}
  \end{table*}
  
  \paragraph{Current-primary foresight interface.}
  Predicted futures condition the planner without replacing current evidence.
  Current and future visual tokens are projected to width 256 (Linear+LayerNorm), then given shared spatial embeddings and learnable per-horizon time embeddings.
  A sample-shared sigmoid gate scales each future stream before fusion; it encodes dataset-level horizon preference, not per-sample reliability.
  Fusion is one Post-LN cross-attention layer (8 heads, no FFN): current tokens are residual queries, and gated futures are keys/values.
  
  Complementary \emph{predicted} future ego motion is injected as planner memory (not ground-truth future status).
  Stop-gradient predicted velocity and acceleration are embedded to width 256, scaled by a second sample-shared sigmoid gate, and concatenated with the measured current ego embedding.
  Navigation commands are not routed through this path.
  The resulting ego memory conditions every decoder layer with the fused visual tokens.

\paragraph{Trajectory-Adaptive Bias.}
TAB is a soft geometry prior that links BEV trajectory candidates to front-view image tokens.
It provides a trajectory-conditioned spatial bias rather than learned sample-specific attention weights.
Each candidate pose is projected into the camera.
A pose is valid only if camera depth exceeds \(10^{-3}\) and the pixel lies inside the image (no clamping); invalid poses are excluded.
For each mode--token pair, we take the maximum Gaussian affinity (bandwidth \(0.25\)) between the token center and the mode's valid projected waypoints, then convert it to a log-bias with floor \(10^{-6}\).
If a mode has no valid projection, the bias is a uniform shift and leaves the softmax unchanged.

We add this bias to trajectory-to-visual attention logits before softmax in every layer of both DiT stages, broadcast across heads.
Ego-memory cross-attention is unchanged.
Within one DDIM step, the first stage shares the initial noisy-anchor plan for projection; later layers then use that stage's supervised plan.

\paragraph{Truncated diffusion planner.}
Following DiffusionDrive, we use truncated diffusion over 20 fixed \(k\)-means \((x,y)\) anchors; heading is \(\tanh(\cdot)\cdot\pi\).
Training uses a 1{,}000-step DDIM schedule with sample prediction and truncated noise levels from \(\{0,\ldots,49\}\).
Noise is added to normalized anchors (not ground-truth trajectories) and denormalized before decoding.
Each of two cascaded stages is a 5-layer DiT with independently cloned weights.
A stage predicts a clean plan by residual xy update plus a heading head.
Training draws one truncated noise sample and runs both stages in one forward.
At the stage boundary, first-stage xy is stop-gradient-copied at the same noise level without re-noising.
Sinusoidal timestep embeddings pass through a small MLP for AdaLN-style modulation.

Mode assignment uses hard closest-anchor matching on mean-horizon xy \(L_2\).
Classification uses sigmoid focal loss (\(\gamma{=}2\), \(\alpha{=}0.25\)); \(L_1\) regression applies only to the winning mode, including heading.
At inference we run two truncated DDIM steps starting near \(t{\approx}8\), execute the full multi-stage decoder each step, and select the mode by classification \(\arg\max\), without an external scorer.

\begin{table}[t]
  \centering
  \setlength{\tabcolsep}{1.5mm}
  \begin{tabular}{@{}llcc@{}}
  \toprule
  Horizons (s) & Type & w/o TAB$\uparrow$ & w/ TAB$\uparrow$ \\
  \midrule
  $\{1\}$ & near-term & 89.1 & 89.6 \\
  $\{2\}$ & mid-term & 89.3 & 89.8 \\
  $\{1,2\}$ & two-step & 89.3 & 89.7 \\
  $\{1,2,3,4\}$ & four-step & 89.6 & 89.9 \\
  \bottomrule
  \end{tabular}
  \caption{Future-query sets vs.\ PDMS (NAVSIM). Default Full: \(1/2/3/4\,\mathrm{s}\) (\(\{1,3,5,7\}\) at 2\,Hz).}
  \label{tab:horizon}
\end{table}

\begin{figure*}[t]
  \centering
  \includegraphics[width=\textwidth]{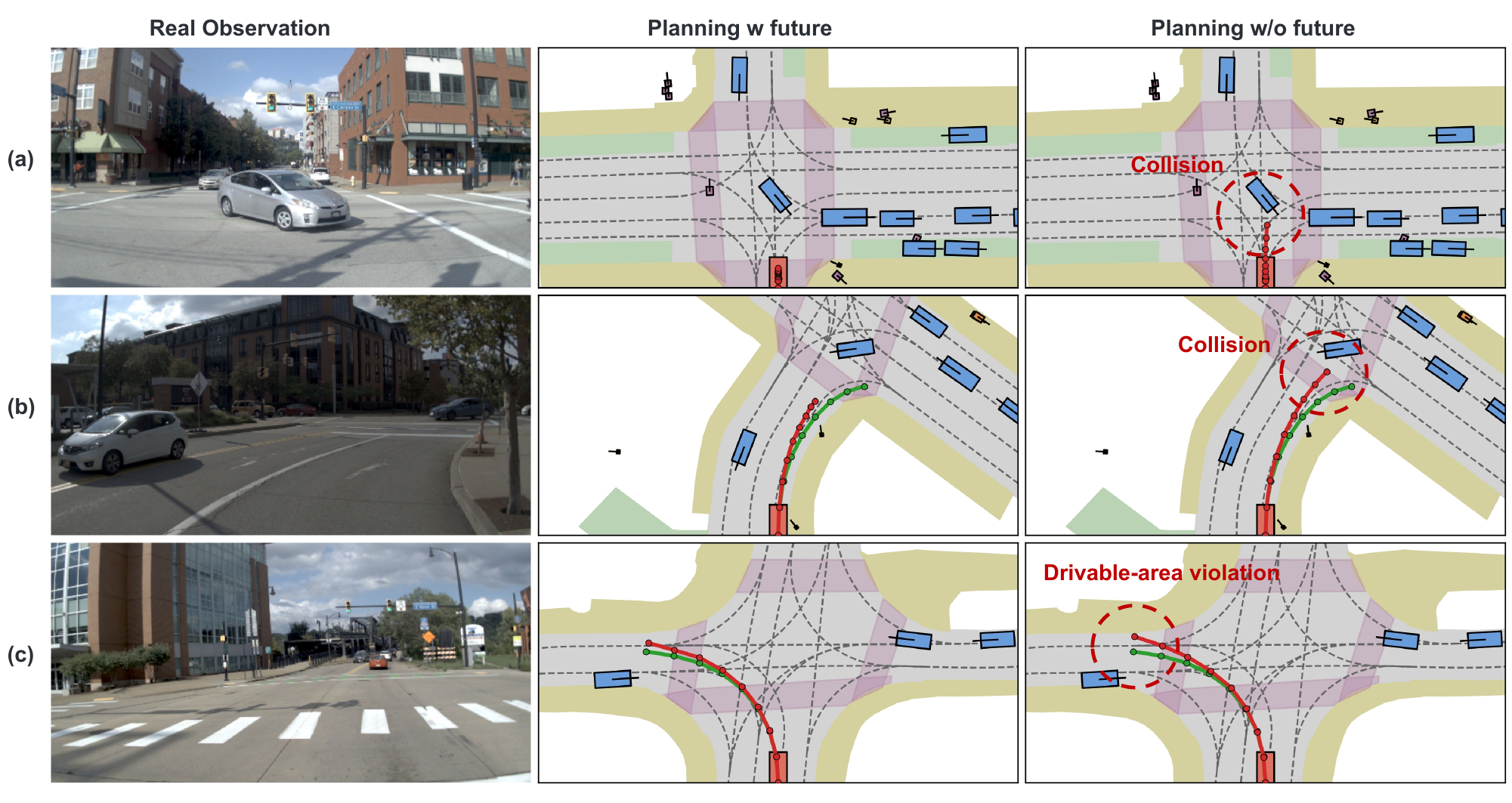}
  \caption{Qualitative comparison only: Full (WM+TAB+status) vs.\ Base+TAB (no WM; matched DiT). This isolates the practical effect of future conditioning rather than a controlled ablation. Left: front view; middle/right: BEV plans (green: expert; red: prediction; blue: agents). Base+TAB fails via collisions~(a,b) or DAC~(c) with per-scene PDMS \(0\); Full remains feasible.}
  \label{fig:wm-compare}
\end{figure*}

\paragraph{Joint optimization.}
We train for 100 epochs with AdamW (base lr \(6\times10^{-4}\), weight decay \(10^{-4}\), cosine decay) on 32 GPUs, global batch 1{,}024, and gradient clipping \(1.0\).
Norm and bias parameters receive no weight decay.
EMA decay for the target encoder is \(0.999\).
Joint loss weights for classification / regression / trajectory / latent / status are \(10 / 8 / 12 / 5 / 0.1\).
Planning gradients update the shared online encoder but stop before the latent predictor, so injected futures are consumed as a stop-gradient signal (main-paper asymmetric recipe; matched inject ablation in Sec.~\ref{app:detach-vs-joint}).
The stop-gradient is applied only on the injected future representations before planner consumption; gradients from planning still update the online encoder through the current representation path.

Backbone fine-tuning depends on encoder capacity.
For DINOv3 ViT-S/16 (ViT-S Base decomposition and small-encoder ablations), we use \emph{full fine-tuning}: layer-wise decay \(1.0\), no stochastic depth, three-epoch warmup, and the same AdamW groups as the heads.
The small encoder can adapt directly to driving with little extra regularization.

For default DINOv3 ViT-B/16, we use a conservative \emph{layered} schedule.
Layer-wise decay \(0.7\) shrinks early-block learning rates; the predictor, fusion modules, and DiT keep the full base rate.
We also use stochastic depth \(0.15\), five-epoch warmup, and BF16 in the backbone.
ViT-L capacity runs use the same schedule except layer-wise decay \(0.8\).
Early pretrained features are largely preserved; adaptation concentrates in upper encoder layers and the WM/planning heads.
DropPath widens the online--EMA gap in training and is disabled at evaluation, so it is a training regularizer rather than a deployment mismatch.

\paragraph{Compute and software.}
\label{app:compute}
We train with PyTorch~2.4 and PyTorch Lightning~2.2 on a standard NAVSIM stack (Python~3.9, CUDA~12.1).
Training uses \textbf{32$\times$NVIDIA H20} GPUs (96\,GB) as \textbf{4 nodes$\times$8 GPUs} with DDP and BF16; per-GPU batch size is 32 (global batch 1{,}024).
Randomness is controlled by Lightning \texttt{seed\_everything}.
The reported ViT-L capacity checkpoint uses seed~0; the default ViT-B main-table score reports a primary run plus three-seed stability, as in the main paper.
Inference latency is measured on one idle H20 with batch size~1 and \(256\times512\) inputs as pure FP32 \texttt{forward} time (excluding crop/resize and data loading).
Training uses the official NAVSIM \texttt{navtrain} latent cache and OpenScene v1.1; evaluation follows official PDMS / one-stage EPDMS.

\section{Baseline Decomposition}
\label{app:base-decomp}

Table~\ref{tab:base-decomp} constructs the current-only \textbf{ViT-S Base} used in small-encoder ablations (NAVSIM v1, DINOv3 ViT-S/16).
DiffusionDrive (88.1 PDMS) is an external reference only: it uses camera+LiDAR BEV and is not a controlled drop-in for our camera-only setting.
Starting from the same truncated-diffusion planner, we replace the BEV interface with a single front-view DINOv3 stream and drop LiDAR~(a), lowering PDMS to 86.4.
Present-status FiLM~(b) adds \(+0.2\); deepening each DiT stage from 1 to 5 layers~(c) yields ViT-S Base at 87.4.

ViT-S Base is thus a current-only camera planner: present-status FiLM and a deeper DiT, but no world model and no TAB.
Present-status FiLM is not the full model's predicted future-status injection.
Steps~(a)--(c) jointly change sensors and representation, so they are uncontrolled adaptations; controlled WM/TAB gains appear in Table~\ref{tab:wm-tab-factorial-vits}.
The main-paper Base uses ViT-B.

\begin{table}[t]
  \centering
  \setlength{\tabcolsep}{1mm}
  \begin{tabular}{@{}lcccc@{}}
  \toprule
  Setting & WM & TAB & PDMS$\uparrow$ & $\Delta$ vs.\ Base \\
  \midrule
  Base &  &  & 87.4 & - \\
  Base + TAB &  & \(\checkmark\) & 88.1 & +0.7 \\
  Base + WM & \(\checkmark\) &  & 88.0 & +0.6 \\
  Base + WM + TAB & \(\checkmark\) & \(\checkmark\) & \textbf{89.0} & +1.6 \\
  \bottomrule
  \end{tabular}
  \caption{WM$\times$TAB factorial on DINOv3 ViT-S/16 (mirrors the main paper). $\Delta$: PDMS vs.\ Base. WM includes aux.\ prediction and visual/status injection.}
  \label{tab:wm-tab-factorial-vits}
\end{table}

\begin{table}[t]
  \centering
  \setlength{\tabcolsep}{0.8mm}
  \begin{tabular}{@{}lccccc@{}}
  \toprule
  Setting & Aux. & \(z\) Inject. & \(s\) Inject. & PDMS$\uparrow$ & $\Delta$ vs.\ Base \\
  \midrule
  Base &  &  &  & 87.4 & - \\
  Auxiliary & \(\checkmark\) &  &  & 87.7 & +0.3 \\
  Visual only & \(\checkmark\) & \(\checkmark\) &  & 87.8 & +0.4 \\
  Status only & \(\checkmark\) &  & \(\checkmark\) & 87.9 & +0.5 \\
  Full WM & \(\checkmark\) & \(\checkmark\) & \(\checkmark\) & 88.0 & +0.6 \\
  \bottomrule
  \end{tabular}
  \caption{WM components without TAB on DINOv3 ViT-S/16 (mirrors the main paper). \(z\)/\(s\): visual/status injection; stop-gradient by default.}
  \label{tab:ablation-vits}
\end{table}

\begin{table}[t]
  \centering
  \setlength{\tabcolsep}{1mm}
  \begin{tabular}{@{}lcccc@{}}
  \toprule
  Setting & WM & TAB & PDMS$\uparrow$ & $\Delta$ vs.\ Base \\
  \midrule
  Base &  &  & 89.3 & - \\
  Base + TAB &  & \(\checkmark\) & 90.0 & +0.7 \\
  Base + WM & \(\checkmark\) &  & 89.8 & +0.5 \\
  Base + WM + TAB & \(\checkmark\) & \(\checkmark\) & \textbf{90.4} & +1.1 \\
  \bottomrule
  \end{tabular}
  \caption{WM$\times$TAB factorial on DINOv3 ViT-L/16 (mirrors the main paper). Full matches the main encoder-capacity result.}
  \label{tab:wm-tab-factorial-vitl}
\end{table}

\begin{table}[t]
  \centering
  \setlength{\tabcolsep}{0.8mm}
  \begin{tabular}{@{}lccccc@{}}
  \toprule
  Setting & Aux. & \(z\) Inject. & \(s\) Inject. & PDMS$\uparrow$ & $\Delta$ vs.\ Base \\
  \midrule
  Base &  &  &  & 89.3 & - \\
  Auxiliary & \(\checkmark\) &  &  & 89.4 & +0.1 \\
  Visual only & \(\checkmark\) & \(\checkmark\) &  & 89.5 & +0.2 \\
  Status only & \(\checkmark\) &  & \(\checkmark\) & 89.6 & +0.3 \\
  Full WM & \(\checkmark\) & \(\checkmark\) & \(\checkmark\) & 89.8 & +0.5 \\
  \bottomrule
  \end{tabular}
  \caption{WM components without TAB on DINOv3 ViT-L/16 (mirrors the main paper). \(z\)/\(s\): visual/status injection; stop-gradient by default.}
  \label{tab:ablation-vitl}
\end{table}

\section{Base vs.\ Base+TAB Attention}
\label{app:uv-attention}

We next ask what \textbf{TAB} changes in planner spatial attention on the main-paper ViT-B Base (current-only; no WM).
TAB should pull trajectory-to-image attention toward the projected plan---typically the near road---rather than a near-uniform front-view scan.
We isolate this effect without a world model: two ViT-B Base runs share data and decoder and differ only in TAB bandwidth (\(\sigma{=}0\) vs.\ \(\sigma{=}0.25\)).
This is not the ViT-S Base of Sec.~\ref{app:base-decomp}; the goal is mechanism visualization, not cross-scale score comparison.

In Fig.~\ref{fig:uv-attention}, Base attention is diffuse over the lower image, while Base+TAB collapses onto a near-road band along the plan.
TAB therefore reshapes spatial evidence gathering, not only the trajectory head prior.
The direction matches the matched Base\(\to\)Base+TAB PDMS gain; causal attribution still rests on the main-paper TAB ablations.

\section{Future-Latent Causal Interventions}
\label{app:future-causal}

Does a WM-conditioned planner use predicted future latents, or only a generic ``non-zero'' signal?
We fix a checkpoint and change only the future tokens at inference.
We report PDMS on NAVSIM \texttt{navtest} (\(N{=}12{,}147\)).
Modes: \textbf{normal} (WM predictions); \textbf{oracle} (EMA ground-truth futures); \textbf{horizon\_shuffle} (permute horizons, keep slot-wise time embeddings); \textbf{token\_mask} (drop \(75\%\) of future tokens); \textbf{cross\_sample} (swap another sample's predictions); \textbf{persistence} (tile the current latent); \textbf{zero} (remove futures).

\paragraph{Isolating the visual future channel.}
We intervene on a checkpoint trained without TAB and without predicted future-status injection (gated fusion only; Table~\ref{tab:app-causal-no-crutch}).
Degradations must then come from the visual future channel.
\textbf{Normal} leads among predicted-future settings: fusion sees in-distribution WM tokens.
\textbf{Oracle} gains only \(+0.2\): GT futures are cleaner but still mismatched to the predictor-trained fusion path.
\textbf{Horizon shuffle} matches normal on every subscore (\(89.4\)).
With the main-paper permutation result, this suggests the DiT uses multi-horizon futures as aggregated multi-scale temporal context rather than a strict ordered rollout; over-smoothed predictions further weaken the order probe.
\textbf{Token mask} (\(-0.4\)) is mild: remaining in-distribution tokens still help.
The key contrast is \textbf{zero} versus identity corruptions.
Zeroing drops PDMS by \(1.4\), mainly via EP/DAC (EP \(83.6{\to}81.2\), DAC \(97.1{\to}95.8\)), while TTC/NC slightly improve.
Without futures the policy is more conservative but less progressive---bad for PDMS, not mainly via collisions.
\textbf{Persistence} and \textbf{cross-sample} fall below zero (\(87.0\) vs.\ \(88.0\)): gated fusion can down-weight near-empty inputs, yet still trusts structured but semantically wrong futures; TTC collapses (\(93.7\) / \(93.2\)) despite higher EP than zero.
Summary: matched predictions help; GT helps little under mismatch; missing futures hurt progress; \emph{wrong} futures hurt collision metrics more than no futures.
A remaining limitation is that predicted latents are over-smoothed relative to GT.

\section{Qualitative Future Conditioning}
\label{app:wm-compare}

\begin{figure}[t]
  \centering
  \includegraphics[width=\linewidth]{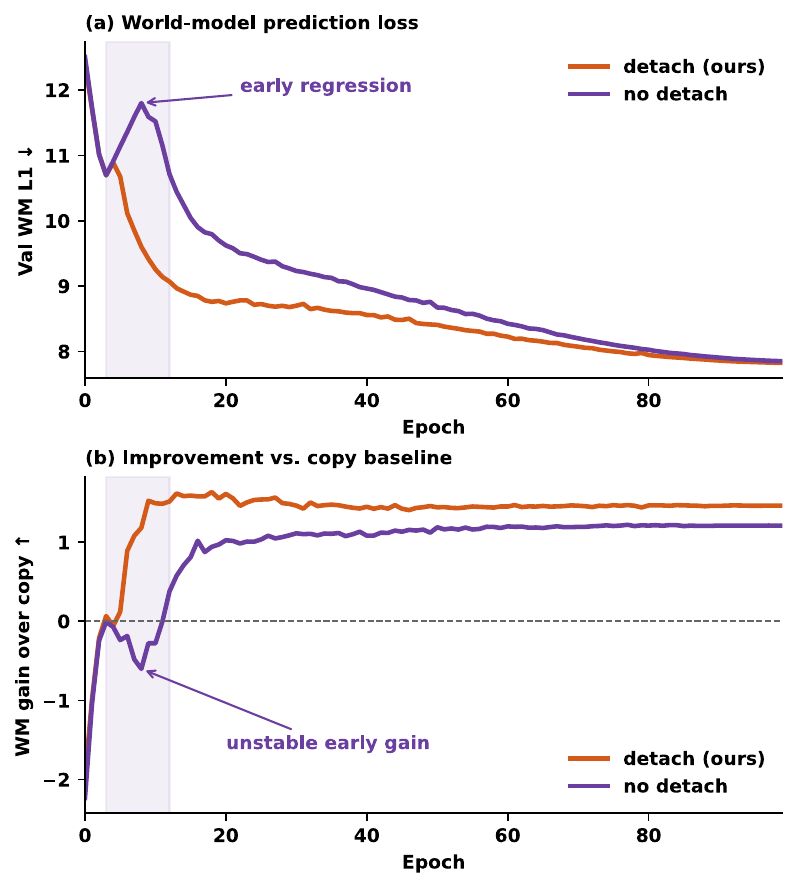}
  \caption{Early WM probes under matched inject (ViT-B; TAB off). (a)~Val.\ WM \(L_1\). (b)~Gain over copy, \(L_1^{\mathrm{copy}}-L_1^{\mathrm{WM}}\) (\(>0\): better than pasting the current latent). Joint shows an early rebound; detach is monotonic. Primary planning claim: Table~\ref{tab:detach-vs-joint}.}
  \label{fig:detach-vs-joint-early}
\end{figure}

\begin{figure*}[t]
  \centering
  \includegraphics[width=\textwidth]{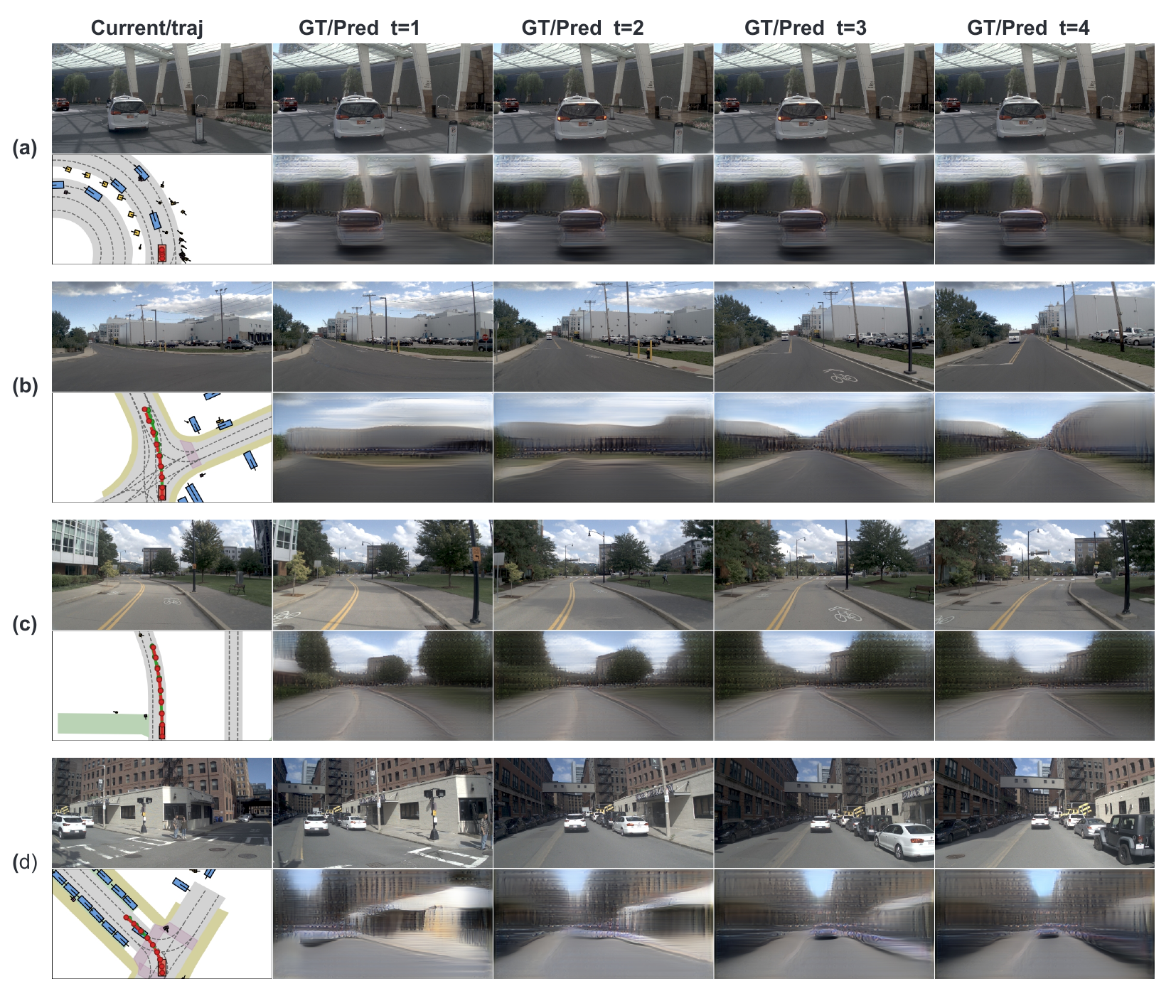}
  \caption{Latent RGB probe. Left: current view and BEV plan. Right: GT frames vs.\ reconstructions from predicted futures (\(t{=}1\)--\(4\); probe trained on EMA targets only). Coarse but scene-grounded; not used for planning.}
  \label{fig:rgb-probe}
\end{figure*}

Figure~\ref{fig:wm-compare} qualitatively compares \textbf{Full} (WM+TAB+status) with \textbf{Base+TAB} (TAB on; no WM / no future injection) under a matched DiT backbone.
Unlike Sec.~\ref{app:future-causal} (TAB/status off), this isolates the practical effect of future conditioning rather than a controlled ablation, and is not the same protocol as Table~\ref{tab:app-causal-no-crutch}.

In the multi-agent intersection~(a) and turning conflict~(b), Base+TAB cuts across an interacting vehicle (per-scene PDMS \(0\); collisions), while Full stays clear (1.00 / 0.58).
In the left-turn corridor~(c), Base+TAB leaves the drivable area (per-scene PDMS \(0\)); Full tracks the expert (1.00).
The cases illustrate two failure modes that foresight can mitigate: dynamic collisions and static DAC errors when a TAB-equipped, future-free decoder misreads the corridor.
They are illustrative; aggregate evidence is in the quantitative tables.

\section{Predicted-Horizon Sets}
\label{app:horizon}

Causal interventions leave open which future-query sets to train with.
Because horizon shuffle leaves PDMS unchanged (Sec.~\ref{app:future-causal}), we treat horizons as complementary multi-scale context, not a strict ordered rollout.
Table~\ref{tab:horizon} compares query sets under matched training, with and without TAB.
Default Full uses \(1/2/3/4\,\mathrm{s}\).

Without TAB, four horizons reach 89.6 PDMS vs.\ 89.1 for \(\{1\}\) and 89.3 for \(\{2\}\) and \(\{1,2\}\).
With TAB, all settings improve and the range shrinks from 0.5 to 0.3; \(\{1,2,3,4\}\) remains best at 89.9.
Aggregating scales helps, but single-run gaps are small: we do not claim any single horizon is necessary, nor that inference requires strict order.
Learned gates likewise do not attribute the gaps to specific horizons.

\section{Encoder-Scale Ablations (ViT-S / ViT-L)}
\label{app:encoder-ablations}

We repeat the main-paper WM$\times$TAB factorial and the WM-component ablation (TAB off) at DINOv3 ViT-S/16 and ViT-L/16.
Protocol matches the main tables (NAVSIM v1 PDMS; single-run).
Full ForeDrive at ViT-S/L matches the main encoder-capacity results (89.0 / 90.4); ViT-S Base matches Table~\ref{tab:base-decomp} (87.4).
With the main-paper ViT-B results (Base 88.9 \(\rightarrow\) Full 89.9), these tables test whether the mechanisms are capacity artifacts.

\paragraph{WM$\times$TAB factorial.}
On ViT-S (Table~\ref{tab:wm-tab-factorial-vits}), Base is 87.4.
TAB / WM alone reach 88.1 / 88.0 (\(+0.7\) / \(+0.6\)); combining both reaches 89.0 (\(+1.6\)).
The joint gain exceeds either factor and the ViT-B combined gain (\(+1.0\)); this single run is suggestive, not conclusive, of a larger weak-encoder benefit.
On ViT-L (Table~\ref{tab:wm-tab-factorial-vitl}), Base is 89.3.
TAB / WM alone add \(+0.7\) / \(+0.5\); Full reaches 90.4 (\(+1.1\)).
TAB adds \(0.7\) at both scales; WM-only gains are \(0.6\) and \(0.5\).
Gains are positive at both scales; attributing the WM-gap to capacity needs more runs.
WM and TAB remain complementary: Full beats the better single factor by \(+0.9\) (ViT-S) and \(+0.4\) (ViT-L).

\paragraph{WM-component ablation (TAB off).}
Tables~\ref{tab:ablation-vits} and~\ref{tab:ablation-vitl} isolate auxiliary prediction vs.\ visual/status injection.
On ViT-S, aux / visual / status / Full WM add \(+0.3\) / \(+0.4\) / \(+0.5\) / \(+0.6\).
On ViT-L, the corresponding gains are \(+0.1\) / \(+0.2\) / \(+0.3\) / \(+0.5\).
Across encoders, direct injection beats auxiliary prediction alone.

\paragraph{Takeaway across scales.}
Main-paper encoder-capacity already summarizes Base\(\to\)Full (\(+1.6\) / \(+1.0\) / \(+1.1\) at ViT-S/B/L).
The tables above give the single-factor and WM-component breakdowns; each tested scale improves in the single-run setting.

\section{Stop-Gradient Routing for Future Latent Injection}
\label{app:detach-vs-joint}

The main paper shows that joint encoder updates help planning, but leaves a finer inject-path choice: should \(\mathcal L_{\rm plan}\) also flow into the world-model predictor?
We isolate that flag under matched future injection.
We change only whether injected visual latents and predicted future status are stop-gradient before the planner.
All else is matched: ViT-B/16, predictor width 512, \(16{\times}32\) tokens, gated fusion, predicted status injection on, TAB off (\(\sigma_{\rm TAB}{=}0\)).
The stop-gradient is applied only on the injected future representations before planner consumption; gradients from planning still update the online encoder through the current representation path.

\begin{table}[t]
  \centering
  \setlength{\tabcolsep}{1.2mm}
  \begin{tabular}{@{}lcccc@{}}
  \toprule
  Setting
  & \(L_1\)\(\downarrow\)
  & Cos.\(\uparrow\)
  & \(\mathcal{L}_{s}\)\(\downarrow\)
  & PDMS$\uparrow$ \\
  \midrule
  Inject + detach (ours)
  & \textbf{8.06}
  & 0.743
  & 0.532
  & \textbf{89.6} \\
  Inject + joint
  & 8.08
  & \textbf{0.747}
  & \textbf{0.514}
  & 89.2 \\
  \bottomrule
  \end{tabular}
  \caption{Detach vs.\ joint through the inject path (ViT-B; TAB off; matched inject). \(L_1\)/Cos./\(\mathcal{L}_{s}\): WM probes on 18{,}179 val samples from \texttt{last.ckpt}.}
  \label{tab:detach-vs-joint}
\end{table}

\begin{table}[t]
  \centering
  \setlength{\tabcolsep}{1.5mm}
  \begin{tabular}{@{}lccc@{}}
  \toprule
  Source
  & ADE$\downarrow$
  & FDE$\downarrow$
  & Shuffle ADE$\downarrow$ \\
  \midrule
  ForeDrive (pred.)
  & \textbf{1.55}
  & \textbf{2.77}
  & 2.52 \\
  Pred.-opt.\ (pred.)
  & 1.83
  & 3.24
  & 3.38 \\
  EMA-GT
  & 2.10
  & 3.61
  & 2.93 \\
  Current
  & 2.31
  & 4.08
  & -- \\
  \bottomrule
  \end{tabular}
  \caption{Frozen-backbone trajectory probe: two-layer MLP regresses expert futures from one latent source. Pred.: WM predictions; EMA-GT: EMA targets; Shuffle ADE: horizon permutation at eval.}
  \label{tab:traj-probe}
\end{table}

\paragraph{Why detach.}
Under matched inject, stop-gradient improves PDMS by \(0.4\) (\(89.6\) vs.\ \(89.2\); Table~\ref{tab:detach-vs-joint}), while final WM probes stay tied (\(L_1\) differs by \(0.02\); joint is slightly better on cosine/\(\mathcal{L}_{s}\)).
Detach is not justified by a higher terminal forecast score.
We use it for asymmetric role separation: the WM is trained only by prediction losses, and the planner consumes stop-gradient futures so trajectory supervision does not rewrite the WM objective.
The PDMS gain aligns with early-training stability (Fig.~\ref{fig:detach-vs-joint-early}) and clearer objective separation, not with lower absolute prediction error.
We do not claim a theoretical resolution of gradient interference.

\paragraph{Early-training motivation.}
Figure~\ref{fig:detach-vs-joint-early} shows training dynamics.
Panel~(a): validation WM \(L_1\).
Panel~(b): gain over copy,
\begin{equation}
\mathrm{gain} \;=\; L_1^{\mathrm{copy}} - L_1^{\mathrm{WM}},
\end{equation}
where \(L_1^{\mathrm{copy}}\) pastes the current latent to every future horizon.
Positive gain means the WM beats this baseline.
In epochs~3--8, joint shows a val.\ \(L_1\) rebound (\(10.69{\to}11.80\)); detach decreases monotonically.
Detach turns stably positive earlier; joint oscillates near zero longer.
After this phase both improve, and final WM probes nearly align (Table~\ref{tab:detach-vs-joint}).
We treat the early curves as an engineering motivation for stop-gradient---joint can be unstable while the predictor adapts---not as evidence of a better terminal forecast.
The planning conclusion remains: under matched inject and matched final WM quality, detach improves PDMS.

\section{Trajectory Probe of Future Latents}
\label{app:traj-probe}

Detach improves PDMS while final WM reconstruction stays nearly unchanged.
Do predicted latents preserve more trajectory-relevant information, or are they only equally reconstructible?
Closed-loop scores and WM \(L_1\)/cosine cannot separate those cases.
We freeze the vision encoder and world model and train a weak readout to regress expert trajectories from one latent source at a time.
Unlike reconstruction probes, this tests whether latents retain information useful for trajectory regression; it does not prove that the planner must rely on these latents, nor that the representation is planning-optimal.
It complements Sec.~\ref{app:future-causal}: swapping EMA-GT into a predictor-trained planner yields little PDMS gain (mismatch); a freshly trained shallow head can still ask which frozen latent is more trajectory-readable.

\paragraph{Protocol.}
Scenes and readout architecture are fixed; only the latent source changes.
Each condition trains its own two-layer MLP.
Per-horizon latents are spatially mean-pooled and concatenated in time; the backbone stays frozen.
We use about 20k/4k train/val scenes, train 50 epochs, and report best ADE.
As a diagnostic, we permute future horizons at evaluation: a large ADE rise means the readout uses cross-horizon structure.
Sources: (i)~current-frame features; (ii)~ForeDrive predicted futures; (iii)~prediction-optimized predicted futures (better WM reconstruction, weaker PDMS in the main paper); (iv)~EMA target futures.

\paragraph{Results.}
ForeDrive yields the lowest readout error (ADE \(1.55\) vs.\ \(1.83\) / \(2.10\) / \(2.31\); Table~\ref{tab:traj-probe}).
Current is worst, so gains are not current-only leakage.
ForeDrive beats the prediction-optimized future by about \(0.28\) ADE (\({\approx}15\%\)).
The latter matches EMA targets more closely yet scores lower PDMS: better reconstruction need not mean a more actionable future.
ForeDrive also beats EMA-GT under this shallow probe, consistent with planning-joint training favoring trajectory-readable futures over scene-target fidelity.
Horizon shuffle raises ADE by about \(0.8\)--\(1.5\) for all future sources, so latents retain cross-horizon structure.
This contrasts with the planner's near-invariance to shuffle (Sec.~\ref{app:future-causal}): the DiT aggregates multi-scale context robustly, while the latent still preserves temporal structure for an ordered readout.

In short, under a frozen backbone and shallow readout, ForeDrive predicted futures regress expert plans more accurately than a more reconstruction-accurate predicted future, EMA targets, and current-only features.
This provides evidence that predicted latents preserve trajectory-relevant information, rather than serving only as ordinary future-feature augmentation; we do not claim a planning-optimal representation.

\section{Latent RGB Probe Visualization}
\label{app:rgb-probe}

The trajectory probe tests planning readability; we also check scene grounding.
We train a lightweight RGB readout on a \emph{frozen} ForeDrive encoder, without updating the world model or planner.

\paragraph{Probe architecture.}
The readout is a lightweight upsampling decoder from frozen DINOv3 ViT-B patch tokens to \(256\times512\) RGB, trained with weighted L1 + LPIPS.
The same decoder is shared across current and future frames.

\paragraph{Training protocol.}
We freeze the jointly trained online/EMA vision encoder and fit only the probe on up to 20k cached training scenes.
Inputs are frozen encodings of the current frame and WM future frames (about \(0.5\)--\(4\,\mathrm{s}\)); targets are camera RGB.
We optimize \(\mathcal L=1.0\cdot\mathrm{L1}+0.1\cdot\mathrm{LPIPS}\) (AlexNet) with AdamW (lr \(3\times10^{-4}\), wd \(10^{-4}\), batch 16) for 8 epochs.
Training uses EMA latents only---not predictor rollouts---so the probe tests whether the latent space itself is RGB-decodable (final train loss \(\approx0.069\)).
At visualization time we decode both EMA targets and predictor rollouts with the same frozen probe.
Predicted latents lie in the same space via the latent prediction objective, so qualitative comparison needs no predictor update.

\paragraph{Qualitative analysis.}
Figure~\ref{fig:rgb-probe} shows four NAVSIM scenes.
Reconstructions are intentionally coarse, as expected for a latent prediction space not trained for photorealism.
Still, the probe recovers scene structure useful for planning diagnostics---road layout~(c), nearby vehicles and buildings~(a,d), and ego-motion-consistent viewpoint change~(b)---with coherent evolution from \(t{=}1\) to \(t{=}4\).
Predicted futures remain scene-grounded after asymmetric joint training; the RGB readout is diagnostic only and never used for planning.

\section{Accuracy--Size Plot Data}
\label{app:fig2-data}

Table~\ref{tab:fig2-data} lists total parameter counts and NAVSIM scores for methods in the main accuracy--size figure.

\begin{table}[t]
  \centering
  \setlength{\tabcolsep}{3.5pt}
  \begin{tabular}{@{}lccc@{}}
  \toprule
  Method & Params (M) & PDMS$\uparrow$ & EPDMS$\uparrow$ \\
  \midrule
  \multicolumn{4}{@{}l}{\textit{Traditional end-to-end methods}} \\
  TransFuser & 55.9 & 84.0 & 76.7 \\
  PRIX & 37 & 87.8 & 84.2 \\
  DiffusionDrive & 60.7 & 88.1 & 84.5 \\
  MeanFuser & 54.6 & 89.0 & 89.5 \\
  DiffRefiner-R34 & 74.8 & 89.4 & 86.2 \\
  DriveSuprim-R34 & 61 & 89.9 & 83.1 \\
  \midrule
  \multicolumn{4}{@{}l}{\textit{World-model and video--action methods}} \\
  Epona & 2500 & 86.2 & 85.1 \\
  DriveVLA-W0 & 7500 & 87.2 & 86.1 \\
  \midrule
  ForeDrive (ours) & 118.4 & 89.9 & 90.0 \\
  \bottomrule
  \end{tabular}
  \caption{Total parameters and NAVSIM scores for the main accuracy--size figure. Same total-parameter definition throughout; self-reports preferred (TransFuser/DiffusionDrive from MeanFuser when needed).}
  \label{tab:fig2-data}
\end{table}

We count all modules in the evaluated model (encoder, WM, planner, and test-time heads).
Training-only EMA encoders are excluded when distinguished (ForeDrive: 118.4M).
Self-reports are preferred; otherwise we use published third-party totals (MeanFuser for TransFuser/DiffusionDrive).
Marker area in the main figure scales with \(\log\) parameters.
Among traditional end-to-end planners (37--75M), ForeDrive ties the best listed PDMS (89.9, with DriveSuprim-R34) and exceeds the strongest EPDMS (MeanFuser, 89.5) by 0.5 at 118.4M.
Relative to Epona (2.5B) and DriveVLA-W0 (7.5B), it improves both aggregates at one to two orders of magnitude smaller scale.